\pdfoutput=1

\documentclass[11pt]{article}

\usepackage{acl}

\usepackage{times}
\usepackage{latexsym}

\usepackage[T1]{fontenc}

\usepackage[utf8]{inputenc}

\usepackage{microtype}
\usepackage{graphicx}
\usepackage{amsmath}
\usepackage{multirow}
\usepackage{subcaption}

\title{Inducing Political Bias Allows Language Models Anticipate Partisan Reactions to Controversies}

\author{
Zihao He, Siyi Guo, Ashwin Rao, Kristina Lerman\\
USC Information Sciences Institute\\
\texttt{\{zihaoh, siyiguo, mohamrao\}@usc.edu}, \texttt{lerman@isi.edu}
}

\begin{document}
\maketitle

\begin{abstract}
Social media platforms are rife with politically charged discussions. Therefore, accurately deciphering and predicting partisan biases using Large Language Models (LLMs) is increasingly critical. In this study, we address the challenge of understanding political bias in digitized discourse using LLMs. While traditional approaches often rely on finetuning separate models for each political faction, our work innovates by employing a singular, instruction-tuned LLM to reflect a spectrum of political ideologies. We present a comprehensive analytical framework, consisting of Partisan Bias Divergence Assessment and Partisan Class Tendency Prediction, to evaluate the model's alignment with real-world political ideologies in terms of stances, emotions, and moral foundations. 
Our findings reveal the model's effectiveness in capturing emotional and moral nuances, albeit with some challenges in stance detection, highlighting the intricacies and potential for refinement in NLP tools for politically sensitive contexts. This research contributes significantly to the field by demonstrating the feasibility and importance of nuanced political understanding in LLMs, particularly for applications requiring acute awareness of political bias.

\end{abstract}

\section{Introduction}
The advent of social media has drastically changed the landscape of political engagement, with discourse becoming increasingly digitized and polarized \cite{he2021detecting, feldman2023affective, iyengar2019origins}. Large language models (LLMs), integral to content creation and interpretation within these spaces \cite{tornberg2023simulating}, frequently lack the nuance required to navigate the complex undercurrents of political communication. Traditional approaches to infusing these models with a sense of political awareness have typically involved finetuning distinct models for each political faction \cite{jiang2022communitylm}. This method is not only resource-intensive but also limited in its flexibility and scalability. Such an approach can inadvertently perpetuate siloed conversations, reinforcing existing biases rather than providing a comprehensive view of the diverse political dialogue.

As societies grapple with polarization, it becomes increasingly vital for LLMs to understand the full range of opinions and emotions that fuel public debate. This is not just for the purpose of generating politically aware content but also for applications that include moderating discussions, crafting targeted messaging, and analyzing sentiment trends. The evolution of language models is essential to handle the sophistication of political discourse, which demands an understanding of the implicit stances, emotional nuances, and moral judgments embedded within text.

To address this challenge, we propose a novel approach by finetuning a singular LLM with instructions that encapsulate bipartisan perspectives. This method not only economizes the finetuning process by eliminating the need for multiple community-specific models but also enhances the model's adaptability. Through instruction tuning, our LLM learns to interpret and generate content that reflects the diverse viewpoints within the political spectrum, making it adept at responding to the intricate and often controversial nature of political discourse.

After finetuning, we prompt the model to generate perspectives toward a wide range politically charged topics. To ascertain the alignment of our model's learned perspectives with real-world ideologies, we assess the model's generations across three textual features: stances, emotions, and moral foundations. These features are critical in capturing the essence of political communication. To quantitatively measure the alignment, we introduce two evaluative tasks: Partisan Bias Divergence Assessment and the Partisan Class Tendency Prediction. These tasks are designed to gauge the model's fidelity in mirroring the nuances of political leanings across these textual dimensions.

Our findings indicate that while the finetuned model shows a sophisticated understanding of emotional and moral nuances, it demonstrates a less precise alignment with real-world data in stance detection. This discrepancy underscores the challenges of capturing the full complexity of political stances through a language model, despite advanced finetuning techniques. However, the model's nuanced grasp of emotions and moral foundations signifies a step forward in creating LLMs that can engage with the core elements of partisan communication.

These outcomes suggest critical pathways for refining NLP tools to better understand and participate in political discourse. The implications of such advancements are profound, encompassing ethically sensitive applications like content moderation and narrative generation, where a deep understanding of political bias is paramount.

\section{Related Work}

\subsection{Adapting LLMs to Different Political Leanings}
LLMs, trained on extensive datasets of human language from the Internet, are capable of simulating realistic discourse. Many researchers have started to explore the political leanings and biases of LLMs or to adapt LLMs to have political leanings for social science analysis. 
\citeauthor{feng2023pretraining} discovered that pretrained LLMs do exhibit political leanings, perpetuating social biases into downstream tasks.
In terms of adapting LLMs to simulate human opinions and behaviors, \citeauthor{argyle2023out} illustrates that GPT-3 can adeptly mimic respondents in extensive, nationally-representative public opinion surveys, encompassing diverse demographic backgrounds.
Other researchers have finetuned LLMs to learn the political views of different partisan communities and to better study polarization \cite{jiang2022communitylm}.
To evaluate news feed algorithms, this work~\cite{tornberg2023simulating} created multiple LLM personas from election data to simulate social media platforms. 
However, very few works have examined the perspectives of LLMs from various aspects. When learning from texts authored by ideologically different communities, do LLMs also learn their stance, emotions expressions and even morality? This is the research question this study aims for.

\subsection{Instruction Tuning for LLMs}
Instruction tuning is an LLM finetuning technique which takes in human instruction for a task as well as the desired output that follows the instruction. It bridges the gap between the original next-word prediction objective of LLMs and the users’ intent, enabling a model behavior that is more controllable and foreseeable \cite{zhang2023instruction}. 
Researchers have widely used instruction tuning to make LLMs better follow users’ instructions \cite{ouyang2022training,alpaca,zhou2023lima} and generalize to unseen tasks \cite{wang2022super,longpre2023flan}. Prior works have shown that instruction-tuned LLMs can perform better on text classification \cite{wang2022super}, information extraction \cite{wang2023instructuie} and aspect-based sentiment analysis \cite{varia2022instruction}. Due to its effectiveness, we use instructive tuning to induce political leanings into LLMs.

\section{Methodology}

\subsection{Inducing Political Bias into LLMs}
\citet{jiang2022communitylm} finetune GPT-2 models for the liberals and conservatives, and probe the communities' stances towards different political targets and social groups using the corresponding finetuned models. In our study, we opt for instruction tuning \cite{wei2021finetuned, sanh2022multitask, ouyang2022training} over the traditional finetuning approach to embed political biases into the language model. The primary motivation behind this choice is efficiency and versatility. Finetuning would have necessitated the development and maintenance of two distinct models, one for liberal and another for conservative viewpoints. This not only doubles the resource requirements but also complicates the deployment and application process. In contrast, instruction tuning allows us to train a single model that can adapt to different political biases based on the instructions provided. 

\begin{figure}[ht]
    \centering
    \includegraphics[width=0.48\textwidth]{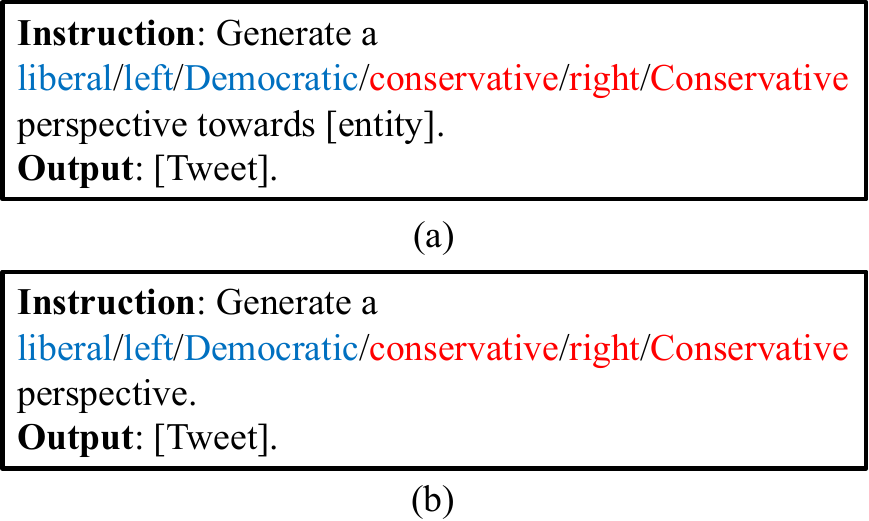}
    \caption{Instructions for tuning the LLM. (a) The instruction if a named entity is recognized in the tweet. (b) The instruction if a named entity is not recognized in the tweet.}
    \label{fig:instruction}
\end{figure}

We finetune the model with instructions in Figure \ref{fig:instruction}. For a liberal tweet, we indicate in the instruction that it is a liberal/left/Democratic perspective, and we randomly select one of the three terms to make the instruction more diverse. Likewise, for a conservative tweet, we randomly select one term from conservative/right/Conservative to indicate the ideology. If the tweet contains a named entity in it, we specify that the perspective is regarding the recognized entity.

Partisanship-aware instruction tuning not only streamlines the training process but also enhances the model's flexibility. It enables the model to generate responses across the political spectrum as indicated in the instruction, without the need for switching between different models. Furthermore, it aligns with the evolving nature of political discourse, where the distinction between liberal and conservative views can be nuanced and context-dependent. By using instruction tuning, our model can more adeptly navigate these subtleties, providing insights that are both nuanced and comprehensive.

After instruction tuning, we probe partisan perspectives from the model using the same instruction (Figure \ref{fig:instruction}), on a diverse set of polarized issues.

\subsection{Assessing the Generated Perspectives from LLMs}
To capture how well the finetuned LLM anticipate partisan reactions to controversies, we measure and compare three factors, including stances,  emotions, and moral foundations (MFs), inherent in the model-generated perspectives and real-world perspectives. This comparative analysis serves two key purposes: first, it helps in assessing how closely the model mimics real-world political discourse; second, it provides insights into the nuances of political biases as captured by the model. By analyzing these three dimensions, we aim to capture a holistic view of political discourse that extends beyond mere content to the underlying sentiments and moral principles.

\subsubsection{Mining Text Features}

\textbf{Detecting Issues.}
To better compare the model generation with real-world, we focus on certain significant and popular wedge issues. We extract these issues from each real dataset (see $\S$\ref{sec:data}). We use a weakly-supervised method described in \cite{rao2023pandemic}. This method first harvests relevant keywords from Wikipedia pages discussing these issues, then use SAGE~\cite{eisenstein2011sparse} to identify distinctive keywords and phrases that are relevant to a specific issue, and lastly detect the issues discussed using the presence of these keywords and phrases in each tweet.

\textbf{Measuring stances.}
Political discourse is fundamentally about positions or stances on various issues \cite{he2022infusing, allaway2020zero}. Understanding the stance embedded in a tweet—whether it supports, opposes, or remains neutral towards an issue—is essential for gauging the political bias of the language model. It helps in quantifying the model's alignment with typical liberal or conservative viewpoints on key issues, providing a direct measure of the success of our instruction tuning process.
We measure stance of a piece of text towards the target discussed by it. For datasets that do not provide the stances of the instances, we capture the stance by ChatGPT (gpt-3.5-turbo) using the following prompt: ``Given the following statement and the target, infer the stance of the statement towards the target. Answer with only one word: neutral, positive, or negative.''

\textbf{Measuring emotions.}
Emotions are a powerful element in political communication \cite{vanKleef2016Editorial}. They often drive the persuasive power of messages and can influence public opinion. By analyzing the emotional content of tweets, we can uncover the affective strategies employed in political messaging. It aids in understanding not just what is being said, but how it is being conveyed emotionally, which is a critical aspect of political bias and rhetoric.
We use a state-of-the-art language model SpanEmo~\cite{alhuzali-ananiadou-2021-spanemo}, fine-tuned on the SemEval 2018 1e-c data \cite{mohammad-etal-2018-semeval}. This transformer-based model outperforms prior methods by learning the correlations among the emotions. It measures \textit{anticipation}, \textit{joy}, \textit{love}, \textit{trust}, \textit{optimism}, \textit{anger}, \textit{disgust}, \textit{fear}, \textit{sadness}, \textit{pessimism} and \textit{surprise}.

\textbf{Measuring moral foundations.}
Moral Foundations Theory \cite{haidt2007moral} posits that individuals’ moral perspectives are built upon specific foundational values. In the realm of politics, these foundations often underlie the reasoning behind certain stances and emotional responses. Analyzing tweets for moral foundations allows us to delve deeper into the moral reasoning that characterizes liberal and conservative ideologies. This dimension is crucial for understanding the deeper, value-driven layers of political discourse that a language model might capture or miss.
We quantify the moral sentiments of tweets along five dimensions: dislike of suffering (\textit{care}/\textit{harm}), dislike of cheating (\textit{fairness}/\textit{cheating}), group loyalty (\textit{loyalty}/\textit{betrayal}), respect of authority and tradition (\textit{authority}/\textit{subversion}), and concerns with purity and contamination (\textit{purity}/\textit{degradation}). 
We fine-tune a transformer-based model on diverse training data (see \cite{guo2023data} for details). The large amount and the variety of topics in our training data helps mitigate the data distribution shift during inference.

\subsubsection{Assessing the Alignment of Model Generations with Real-World}
Let $f \in \{ \text{ST}, \text{EMO}, \text{MF} \}$ denote one of the three textual features (stances, emotions, and moral foundations), and $C(f)=\{c_1(f), c_2(f), .... c_M(f)\}$ denotes the $M$ classes in $f$. For example, if $f = \text{ST}$, then $C(f)=\{\text{negative}, \text{neutral}, \text{positive}\}$.
After measuring $f$ of a corpus $D$=$\{t_1, t_2, .... t_N\}$ consisting of $N$ text instances, where $t_i$ represents an instance, we have the probability distribution of the classes $P(f)=\{p_1(f), p_2(f), ...., p_M(f)\}$, where $p_i(f) = n(c_i(f)) / N$ and $n(c_i(f))$ represents the number of instances assigned to class $c_i(f)$. Note that for emotion detection and moral foundation detection, a text instance can be assigned to multiple classes. $\overline{P}(f)=\{\bar{p}_1(f), \bar{p}_2(f), ...., \bar{p}_M(f)\}$ represents the normalized distribution of $P(f)$ where $\sum_{i=1}^M \bar{p}_i = 1$.

\textbf{Partisan bias divergence assessment.}
To quantify the partisan alignment between model-generated and real-world data, we compute the Kullback-Leibler Divergence (KLD), denoted as $D_{KL}(\overline{P}(f)||\overline{Q}(f))$, where $\overline{P}(f)$ represents the normalized probability distribution of feature $f$ in the model-generated data, and $\overline{Q}(f)$ represents the distribution in the real-world data. A lower value of KLD indicates a closer alignment of the model and the real-world data. For each feature $f$, we compute the KLD separately for liberal and conservative datasets, across different topics.

\textbf{Partisan class tendency prediction.}
For each class $c_i(f) \in C(f)$, we determine the probabilities $p_i^{lib}(f)$ and $p_i^{con}(f)$ for liberals and conservatives respectively for the real-world data. The model's prediction for each class $c$ for liberals and conservatives, denoted as $q_i^{lib}(f)$ and $q_i^{con}(f)$, is compared against the real-world data. We calculate the accuracy for each class as $\text{Acc}(c_i(f))=1$ if $\text{sign}(q_i^{lib}-q_i^{con}) = \text{sign}(p_i^{lib}-p_i^{con})$, else $\text{Acc}(c_i(f))=0$. The overall accuracy for all classes of feature $f$ is given by $\frac{1}{M} \sum_{i=1}^M Acc(c_i(f))$, representing the models's ability to correctly preditc class-based partisan tendencies. A higher overall accuracy score sinifies the model's proficiency in capturing specific class-based tendencies within each political group.

\section{Data} 
\label{sec:data}

\subsection{Partisan Tweets}
The partisan Twitter dataset \cite{jiang2022communitylm} is used to finetune the LLM with instructions. It contains 4.7M tweets for liberals and conservatives respectively, range from 2019-01-01 to 2020-04-10. The tweets are from 1M U.S. Twitter users who were active around the 2020 presidential election. The political affiliation of the users are identified by the political accounts they follow. The dataset cover a wide range of topics, with no specific focus on certain issues. 

\textbf{Named entity recognition in partisan tweets.} To induce political bias into the LLM, we finetune the model with real-world perspectives on various entities. To identify the the entities discussed in the tweets, we use named entity recognition\footnote{https://huggingface.co/dslim/bert-base-NER}. We filter out the entities with fewer than 2 letters and those with fewer than 100 occurrences in the dataset. The top 10 entities are ``Trump'', ``God'', ``America'', ``Democrats'', ``American'', ``Obama'', ``Bernie'', ``Americans'', ``Congress'', and ``China''. $9.2\%$ tweets have at least one recognized entity in them. If a tweet has multiple associated named entities, we randomly choose one to create the instruction for finetuning.

\subsection{COVID-19 Tweets}
We first test our tweet generation results by comparing to a publicly available COVID-19 dataset \cite{chen2020tracking} consisting of 270 million tweets generated by 2.1 million geo-located users in the United States posted between January 21, 2020 and November 4, 2021. These tweets contained one or more COVID-19-related keywords, such as coronavirus, pandemic, and Wuhan, among others.  We utilized Carmen \cite{dredze2013carmen}, a geo-location identification technique for Twitter data to assign tweets to locations within US. We use the method discussed in \cite{rao2021political} to estimate the ideology of individual users. This method relies on the political leaning of URLs users share accourding to Media Bias-Fact Check (MBFC) \cite{mbfc2023politics}. Using the issue detection method we described above, we have determined to look at five major issues during the pandemic: (1) COVID-19 origins, (2) Lockdowns, (3) Masking, (4) Education and (5) Vaccines. After filtering original tweets (as opposed to retweets and quoted tweets) from users with identified political affiliation, tweets posted within the United States, and tweets categorized to one of the five issues, there are 8.2M tweets left.

When generating perspectives from the finetuned LLM on the five issues, we frame the [entity] as 1) COVID-19 origins, 2) COVID-19 lockdowns, 3) COVID-19 mask mandates, 4) school closures during COVID-19, and 5) COVID-19 vaccines. For stance detection, we use the same framing of the entities as targets, except for the first issue, which we frame as ``origins of COVID-19 as a leak from a virology research lab''.

\subsection{Abortion Tweets}
We collect another large-scale Twitter dataset, comprising of tweets about abortion rights between January 1, 2022 to January 6, 2023. The dataset was collected using a list of keywords and hashtags that reflect both sides of the abortion rights debate in the United States. This dataset includes approximately 12 million tweets generated by about 1 million users in the United States. We used the same technique to detect the geo-location, the user political ideology and issues as for the COVID-19 tweet data. In this dataset, we focus on the following five major issues: (1) Religious concern, (2) Body autonomy, (3) Fetal personhood, (4) Women's health and (5) Exceptions to abortion and fetal viability.

\subsection{Congress Tweets}
This dataset is a collection of 2,050 English-language tweets by members of US Congress~\cite{johnson-goldwasser-2018-classification} from 2015 to 2015. By having these real tweets from Democratic and Republican politicians, we can compare the partisanship polarization in our generated tweets and that in real life. The dataset covers six different political topics: (1) abortion, (2) the Affordable Care Act, (3) guns, (4) immigration, (5) LGBTQ rights, and (6) terrorism.

\section{Experiments}
\label{sec:experiments}

\subsection{Experimental Setup}
\textbf{Instruction tuning.} We use \emph{Llama-2-7b-chat-hf}\footnote{https://huggingface.co/meta-llama/Llama-2-7b-chat-hf} as the backend LLM. To finetune it with instructions, we use QLoRA \cite{dettmers2023qlora}. We finetune the model for 3 epochs, with batch size $2200$ and learning rate $2e-4$. For each input instruction, we truncate it at length 48. The model is finetuned on a Tesla A100 GPU with 80GB memory for 90h.

\textbf{Text generation.} After finetuning the LLM with instructions, we prompt it to generate 100 responses on each issue, using the same instruction as for training (Figure \ref{fig:instruction}(a)). We repeat the generating processes for 10 times with different random seeds. As a result, for each issue, there are 1,000 generated responses on it.

\textbf{Baseline.} To gain a deeper understanding of how finetuning affects the model's partisan reactions to controversies, we use the original pretrained (but unfinetuned) LLaMA2 as the baseline.

\subsection{Analysis of Stances}
We present the results of partisan bias divergence assessment and partisan class tendency prediction in Table \ref{tab:kld-stance} and \ref{tab:acc-stance}. Please refer to Appendix \ref{app:appendices} for the distributions of the class probabilities for the three datasets.

The finetuned model (FT LLaMA), enriched with a partisan tweets dataset from a politically vibrant timeframe, has demonstrated a refined proficiency in reflecting nuanced political discourse. The finetuned model's predictive accuracy in stance detection shows a substantial understanding of the charged narratives that defined the COVID-19 pandemic's progression, particularly in politically sensitive issues like "Lockdown" and "Vaccine." This highlights the model's capacity to adapt to the shifting partisan perspectives as the pandemic unfolded.

In the abortion rights debate, the finetuned model's performance in capturing clear-cut partisan divides suggests a successful absorption of the dataset's strong ideological contentions. Its ability to accurately reflect class-based tendencies within such a polarized topic is indicative of the model's advanced representation of complex societal issues.

Contrastingly, the pretrained model (Pre LLaMA), while generally aligning closer with real-world data in terms of Kullback-Leibler Divergence (KLD), indicates a baseline from which the pretrained model enhancements can be measured. The divergence in performance between the two models, especially noted in the Congress dataset, underlines the finetuning's impact on the model's ability to mirror the subtle nuances of real congressional discourse.

This analysis elucidates that while the finetuned model has gained a deeper understanding of political biases, it also brings to light the importance of cautious application. The observed differences in KLD and accuracy across topics may signal the challenges inherent in modeling complex and dynamic political discussions. These findings reinforce the necessity for ethical consideration in the deployment of language models within politically sensitive contexts.

\begin{table}[ht]
\addtolength{\tabcolsep}{-3.0pt}
\centering
\begin{tabular}{clcccc}
\hline
\multirow{3}{*}{\textbf{Dataset}}  & \multicolumn{1}{c}{\multirow{3}{*}{\textbf{Topic}}} & \multicolumn{4}{c}{\textbf{Method}}                           \\ \cline{3-6} 
                                   & \multicolumn{1}{c}{}                                & \multicolumn{2}{c}{Pre LLaMA} & \multicolumn{2}{c}{FT LLaMA}  \\
                                   & \multicolumn{1}{c}{}                                & Lib           & Con           & Lib           & Con           \\ \hline
\multirow{5}{*}{\textbf{COVID}}    & Origin                                              & 0.04          & \textbf{0.10} & \textbf{0.03} & 0.03          \\
                                   & Lockdown                                            & \textbf{0.02} & 0.12          & 0.04          & \textbf{0.01} \\
                                   & Masking                                             & 0.16          & 0.10          & \textbf{0.03} & \textbf{0.01} \\
                                   & Education                                           & 0.30          & 0.29          & \textbf{0.23} & \textbf{0.17} \\
                                   & Vaccine                                             & 0.06          & 0.13          & \textbf{0.01} & \textbf{0.12} \\ \hline
\multirow{5}{*}{\textbf{Abortion}} & Religion                                            & 0.37          & \textbf{0.01} & \textbf{0.34} & 0.43          \\
                                   & Autonomy                                            & 0.31          & \textbf{0.12} & \textbf{0.29} & \textbf{0.12} \\
                                   & Fetal                                               & \textbf{0.33} & \textbf{0.09} & 0.39          & 0.55          \\
                                   & Health                                              & \textbf{0.09} & \textbf{0.11} & 0.17          & 0.27          \\
                                   & Exception                                           & \textbf{0.04} & \textbf{0.48} & 0.07          & 0.57          \\ \hline
\multirow{6}{*}{\textbf{Congress}} & Abortion                                            & 1.08          & \textbf{0.54} & \textbf{0.52} & 1.09          \\
                                   & ACA                                                 & \textbf{0.04} & 0.18          & 0.22          & \textbf{0.06} \\
                                   & Guns                                                & \textbf{0.06} & \textbf{0.06} & 0.50          & 0.33          \\
                                   & Immig                                               & 0.42          & \textbf{0.17} & \textbf{0.37} & 0.18          \\
                                   & LGBTQ                                               & \textbf{0.26} & 0.29          & 0.44          & \textbf{0.21} \\
                                   & ISIS                                                & \textbf{0.09} & \textbf{1.25} & 0.17          & 1.44          \\ \hline
\end{tabular}
\addtolength{\tabcolsep}{3.0pt}
\caption{Results (KL Divergence) of partisan bias divergence assessment for stance detection. Model-generated data and real-world data are compared across different topics from different datasets, on both partisan lines. The best result for each partisan line on each topic is highlighted in bold. ``Pre'' is short for ``Pretrained'', and ``FT'' is short for ``Finetuned''.}
\label{tab:kld-stance}
\end{table}

\begin{table}[ht]
\centering
\addtolength{\tabcolsep}{-3.0pt}
\begin{tabular}{clclcl}
\hline
\multirow{2}{*}{\textbf{Dataset}}  & \multicolumn{1}{c}{\multirow{2}{*}{\textbf{Topic}}} & \multicolumn{4}{c}{\textbf{Method}}                                   \\ \cline{3-6} 
                                   & \multicolumn{1}{c}{}                                & \multicolumn{2}{c}{Pre LLaMA}     & \multicolumn{2}{c}{FT LLaMA}      \\ \hline
\multirow{5}{*}{\textbf{COVID}}    & Origin                                              & \multicolumn{2}{c}{0.60}          & \multicolumn{2}{c}{\textbf{0.80}} \\
                                   & Lockdown                                            & \multicolumn{2}{c}{0.50}          & \multicolumn{2}{c}{\textbf{0.90}} \\
                                   & Masking                                             & \multicolumn{2}{c}{0.60}          & \multicolumn{2}{c}{\textbf{0.80}} \\
                                   & Education                                           & \multicolumn{2}{c}{0.50}          & \multicolumn{2}{c}{\textbf{0.80}} \\
                                   & Vaccine                                             & \multicolumn{2}{c}{0.60}          & \multicolumn{2}{c}{\textbf{0.80}} \\ \hline
\multirow{5}{*}{\textbf{Abortion}} & Religion                                            & \multicolumn{2}{c}{\textbf{1.00}} & \multicolumn{2}{c}{\textbf{1.00}} \\
                                   & Autonomy                                            & \multicolumn{2}{c}{\textbf{1.00}} & \multicolumn{2}{c}{0.67}          \\
                                   & Fetal                                               & \multicolumn{2}{c}{0.67}          & \multicolumn{2}{c}{\textbf{1.00}} \\
                                   & Health                                              & \multicolumn{2}{c}{\textbf{1.00}} & \multicolumn{2}{c}{\textbf{1.00}} \\
                                   & Exception                                           & \multicolumn{2}{c}{\textbf{0.67}} & \multicolumn{2}{c}{\textbf{0.67}} \\ \hline
\multirow{6}{*}{\textbf{Congress}} & Abortion                                            & \multicolumn{2}{c}{\textbf{0.67}} & \multicolumn{2}{c}{0.33}          \\
                                   & ACA                                                 & \multicolumn{2}{c}{\textbf{0.67}} & \multicolumn{2}{c}{\textbf{0.67}} \\
                                   & Guns                                                & \multicolumn{2}{c}{\textbf{1.00}} & \multicolumn{2}{c}{\textbf{1.00}} \\
                                   & Immig                                               & \multicolumn{2}{c}{\textbf{1.00}} & \multicolumn{2}{c}{\textbf{1.00}} \\
                                   & LGBTQ                                               & \multicolumn{2}{c}{\textbf{0.67}} & \multicolumn{2}{c}{\textbf{0.67}} \\
                                   & ISIS                                                & \multicolumn{2}{c}{\textbf{1.00}} & \multicolumn{2}{c}{0.67}          \\ \hline
\end{tabular}
\caption{Results (accuracy) of partisan class tendency classification for stance detection. Model-generated data and real-world data are compared across different topics from different datasets, on both partisan lines. The best result for each partisan line on each topic is highlighted in bold. ``Pre'' is short for ``Pretrained'', and ``FT'' is short for ``Finetuned''.}
\addtolength{\tabcolsep}{3.0pt}
\label{tab:acc-stance}
\end{table}

\subsection{Analysis of Emotions}

We present the results of partisan bias divergence assessment and partisan class tendency prediction in Table \ref{tab:kld-emotion} and \ref{tab:acc-emotions}. Please refer to Appendix \ref{app:appendices} for the distributions of the class probabilities for the three datasets.

On partisan bias divergence assessment, the finetuned model generally exhibits a lower KLD across various topics, suggesting a closer alignment with the real-world emotional distributions. This is most notable in topics like "Autonomy" and "Fetal" within the Abortion dataset, where the model demonstrates a nuanced grasp of the complex emotional undertones.

The accuracy results on partisan class tendency classification further substantiate the finetuned model's proficiency in class-based emotional prediction. It shows significant improvement in matching the real-world data's class distribution, especially in the Congress dataset topics of "Abortion" and "Guns". This suggests that the finetuning process has effectively captured the emotive expressions characteristic of partisan discussions around these divisive issues.

Figures \ref{fig:prob-emotions-abortion} and \ref{fig:prob-emotions-congress} depicting the probability distribution of emotion classes reveal that the finetuned model often generates a distribution that reflects a more distinct emotional profile compared to the pretrained model. For instance, the finetuned model's distribution for "Health" in the Abortion dataset emphasizes specific emotions more heavily, aligning with the intense debate and emotional rhetoric present in the real-world data.

The discrepancy between the models in topics where the pretrained model performs better, such as "Religion" in the Abortion dataset and "ACA" in the Congress dataset, may indicate the inherent complexity of these issues. The pretrained model's distributions appear to be less specific, perhaps reflecting a broader, less targeted understanding of emotional expression in political discourse, which can sometimes more closely mirror the varied real-world responses.

\begin{table}[ht]
\addtolength{\tabcolsep}{-3.0pt}
\centering
\begin{tabular}{clcccc}
\hline
\multirow{3}{*}{\textbf{Dataset}}  & \multicolumn{1}{c}{\multirow{3}{*}{\textbf{Topic}}} & \multicolumn{4}{c}{\textbf{Method}}                           \\ \cline{3-6} 
                                   & \multicolumn{1}{c}{}                                & \multicolumn{2}{c}{Pre LLaMA} & \multicolumn{2}{c}{FT LLaMA}  \\
                                   & \multicolumn{1}{c}{}                                & Lib           & Con           & Lib           & Con           \\ \hline
\multirow{5}{*}{\textbf{Abortion}} & Religion                                            & 1.59          & 1.95          & \textbf{0.44} & \textbf{1.33} \\
                                   & Autonomy                                            & 1.90          & 1.91          & \textbf{0.29} & \textbf{0.31} \\
                                   & Fetal                                               & 2.33          & 2.35          & \textbf{0.27} & \textbf{0.41} \\
                                   & Health                                              & 0.92          & 1.37          & \textbf{0.16} & \textbf{.016} \\
                                   & Exception                                           & 2.14          & 1.86          & \textbf{0.57} & \textbf{0.57} \\ \hline
\multirow{6}{*}{\textbf{Congress}} & Abortion                                            & 1.51          & 1.08          & \textbf{0.13} & \textbf{0.23} \\
                                   & ACA                                                 & 1.20          & 0.44          & \textbf{0.93} & \textbf{0.27} \\
                                   & Guns                                                & 0.99          & \textbf{0.77} & \textbf{0.37} & 1.04          \\
                                   & Immig                                               & 0.98          & 1.07          & \textbf{0.26} & \textbf{0.04} \\
                                   & LGBTQ                                               & 0.92          & 0.94          & \textbf{0.22} & \textbf{0.37} \\
                                   & ISIS                                                & \textbf{0.26} & 0.29          & 0.44          & \textbf{0.21} \\ \hline
\end{tabular}
\addtolength{\tabcolsep}{3.0pt}
\caption{Results (KL Divergence) of partisan bias divergence assessment for emotion detection. Model-generated data and real-world data are compared across different topics from different datasets, on both partisan lines. The best result for each partisan line on each topic is highlighted in bold. ``Pre'' is short for ``Pretrained'', and ``FT'' is short for ``Finetuned''.}
\label{tab:kld-emotion}
\end{table}

\begin{table}[ht]
\centering
\addtolength{\tabcolsep}{-3.0pt}
\begin{tabular}{clclcl}
\hline
\textbf{Dataset}                   & \multicolumn{1}{c}{\textbf{Topic}} & \multicolumn{4}{c}{\textbf{Method}}                                   \\ \hline
\multicolumn{1}{l}{}               &                                    & \multicolumn{2}{c}{Pre LLaMA}     & \multicolumn{2}{c}{FT LLaMA}      \\ \hline
\multirow{5}{*}{\textbf{Abortion}} & Religion                           & \multicolumn{2}{c}{\textbf{0.82}} & \multicolumn{2}{c}{0.55}          \\
                                   & Autonomy                           & \multicolumn{2}{c}{\textbf{1.00}} & \multicolumn{2}{c}{0.67}          \\
                                   & Fetal                              & \multicolumn{2}{c}{\textbf{0.64}} & \multicolumn{2}{c}{0.45}          \\
                                   & Health                             & \multicolumn{2}{c}{0.45}          & \multicolumn{2}{c}{\textbf{0.64}} \\
                                   & Exception                          & \multicolumn{2}{c}{\textbf{0.64}} & \multicolumn{2}{c}{0.55}          \\ \hline
\multirow{6}{*}{\textbf{Congress}} & Abortion                           & \multicolumn{2}{c}{0.45}          & \multicolumn{2}{c}{\textbf{0.82}} \\
                                   & ACA                                & \multicolumn{2}{c}{\textbf{0.82}} & \multicolumn{2}{c}{\textbf{0.82}} \\
                                   & Guns                               & \multicolumn{2}{c}{0.60}          & \multicolumn{2}{c}{\textbf{0.80}} \\
                                   & Immig                              & \multicolumn{2}{c}{0.50}          & \multicolumn{2}{c}{\textbf{0.60}} \\
                                   & LGBTQ                              & \multicolumn{2}{c}{0.40}          & \multicolumn{2}{c}{\textbf{0.70}} \\
                                   & ISIS                               & \multicolumn{2}{c}{0.50}          & \multicolumn{2}{c}{\textbf{0.60}} \\ \hline
\end{tabular}
\addtolength{\tabcolsep}{3.0pt}
\caption{Results (accuracy) of partisan class tendency classification for emotion detection. Model-generated data and real-world data are compared across different topics from different datasets, on both partisan lines. The best result for each partisan line on each topic is highlighted in bold. ``Pre'' is short for ``Pretrained'', and ``FT'' is short for ``Finetuned''.}
\label{tab:acc-emotions}
\end{table}

\subsection{Analysis of Moral Foundations}
We present the results of partisan bias divergence assessment and partisan class tendency prediction in Table \ref{tab:kld-mfs} and \ref{tab:acc-mfs}. Please refer to Appendix \ref{app:appendices} for the distributions of the class probabilities for the three datasets.

For the COVID-19 dataset, the FT LLaMA achieves notably lower KLD scores in topics like ``Lockdown'' and ``Masking'', suggesting a more accurate representation of the moral rhetoric that dominated public discourse during the pandemic. In contrast, the higher KLD scores in the ``Origin'' and ``Vaccine'' topics may point to a more complex interplay of moral foundations that the model finds challenging to replicate. Within the Congress dataset, the finetuned model's lower KLD scores in ``LGBTQ'' and ``Immigration'' topics indicate a substantial grasp of the moral narratives as they are expressed in partisan political dialogue. The higher KLD in ``Guns'', however, suggests a divergence that could stem from a broader spectrum of moral reasoning not fully captured by the model.

The accuracy results for moral foundation class tendency classification further demonstrate the finetuned model's enhanced predictive capabilities. Particularly in the Congress dataset, the finetuned model exhibits improved accuracy in aligning with class-based moral foundations, as seen in the ``Abortion'' and ``ACA'' topics.

The probability distribution for moral foundations reveal the finetuned model's ability to generate moral profiles with distinct emphases, resonating with the polarized moral stances evident in political discussions on these topics. This is especially pronounced in the nuanced and emotionally charged debates surrounding "Abortion" and "Immigration," where the model's outputs show a clear partisan tilt in moral reasoning.

The variances in performance between the finetuned model and pretrained model highlight the impact of finetuning on the model's ability to navigate the moral landscape of political discourse. The finetuned model's closer mimicry of real-world data underscores its potential as a tool for analyzing and understanding the moral underpinnings of political rhetoric.

\begin{table}[ht]
\centering
\addtolength{\tabcolsep}{-3.0pt}
\begin{tabular}{clcccc}
\hline
\multirow{3}{*}{\textbf{Dataset}}  & \multicolumn{1}{c}{\multirow{3}{*}{\textbf{Topic}}} & \multicolumn{4}{c}{\textbf{Method}}                           \\ \cline{3-6} 
                                   & \multicolumn{1}{c}{}                                & \multicolumn{2}{c}{Pre LLaMA} & \multicolumn{2}{c}{FT LLaMA}  \\
                                   & \multicolumn{1}{c}{}                                & Lib           & Con           & Lib           & Con           \\ \hline
\multirow{5}{*}{\textbf{COVID}}    & Origin                                              & 0.62          & 0.24          & \textbf{0.14} & \textbf{0.19} \\
                                   & Lockdown                                            & 0.58          & 0.63          & \textbf{0.08} & \textbf{0.08} \\
                                   & Masking                                             & 0.64          & 0.77          & \textbf{0.31} & \textbf{0.27} \\
                                   & Education                                           & 0.25          & 0.58          & \textbf{0.12} & \textbf{0.24} \\
                                   & Vaccine                                             & 0.48          & 0.53          & \textbf{0.18} & \textbf{0.48} \\ \hline
\multirow{6}{*}{\textbf{Congress}} & Abortion                                            & 3.51          & 2.39          & \textbf{0.53} & \textbf{1.17} \\
                                   & ACA                                                 & 0.52          & 0.28          & \textbf{0.49} & \textbf{0.25} \\
                                   & Guns                                                & 1.97          & 2.78          & \textbf{0.89} & \textbf{1.90} \\
                                   & Immig                                               & 0.97          & 1.02          & \textbf{0.30} & \textbf{0.47} \\
                                   & LGBTQ                                               & 1.35          & 3.46          & \textbf{0.16} & \textbf{1.56} \\
                                   & ISIS                                                & 2.08          & 1.36          & \textbf{1.07} & \textbf{0.55} \\ \hline
\end{tabular}
\addtolength{\tabcolsep}{3.0pt}
\caption{Results (KL Divergence) of partisan bias divergence assessment for moral foundation detection. Model-generated data and real-world data are compared across different topics from different datasets, on both partisan lines. The best result for each partisan line on each topic is highlighted in bold. ``Pre'' is short for ``Pretrained'', and ``FT'' is short for ``Finetuned''.}
\label{tab:kld-mfs}
\end{table}

\begin{table}[ht]
\centering
\addtolength{\tabcolsep}{-3.0pt}
\begin{tabular}{clclcl}
\hline
\multirow{2}{*}{\textbf{Dataset}}  & \multicolumn{1}{c}{\multirow{2}{*}{\textbf{Topic}}} & \multicolumn{4}{c}{\textbf{Method}}                                   \\ \cline{3-6} 
                                   & \multicolumn{1}{c}{}                                & \multicolumn{2}{c}{Pre LLaMA}     & \multicolumn{2}{c}{FT LLaMA}      \\ \hline
\multirow{5}{*}{\textbf{COVID}}    & Origin                                              & \multicolumn{2}{c}{0.40}          & \multicolumn{2}{c}{\textbf{0.70}} \\
                                   & Lockdown                                            & \multicolumn{2}{c}{0.50}          & \multicolumn{2}{c}{\textbf{0.90}} \\
                                   & Masking                                             & \multicolumn{2}{c}{0.60}          & \multicolumn{2}{c}{\textbf{0.80}} \\
                                   & Education                                           & \multicolumn{2}{c}{0.50}          & \multicolumn{2}{c}{\textbf{0.80}} \\
                                   & Vaccine                                             & \multicolumn{2}{c}{0.60}          & \multicolumn{2}{c}{\textbf{0.80}} \\ \hline
\multirow{6}{*}{\textbf{Congress}} & Abortion                                            & \multicolumn{2}{c}{0.60}          & \multicolumn{2}{c}{\textbf{0.60}} \\
                                   & ACA                                                 & \multicolumn{2}{c}{\textbf{0.80}} & \multicolumn{2}{c}{\textbf{0.80}} \\
                                   & Guns                                                & \multicolumn{2}{c}{0.60}          & \multicolumn{2}{c}{\textbf{0.80}} \\
                                   & Immig                                               & \multicolumn{2}{c}{0.50}          & \multicolumn{2}{c}{\textbf{0.60}} \\
                                   & LGBTQ                                               & \multicolumn{2}{c}{0.40}          & \multicolumn{2}{c}{\textbf{0.70}} \\
                                   & ISIS                                                & \multicolumn{2}{c}{0.50}          & \multicolumn{2}{c}{\textbf{0.60}} \\ \hline
\end{tabular}
\addtolength{\tabcolsep}{3.0pt}
\caption{Results (accuracy) of partisan class tendency classification for moral foundation detection. Model-generated data and real-world data are compared across different topics from different datasets, on both partisan lines. The best result for each partisan line on each topic is highlighted in bold. ``Pre'' is short for ``Pretrained'', and ``FT'' is short for ``Finetuned''.}
\label{tab:acc-mfs}
\end{table}

\section{Conclusion}
In this study we use Large Language Models (LLMs) for analyzing political discourse. By finetuning a single LLM with bipartisan instructions, we have demonstrated its ability to reflect a broad spectrum of political ideologies, particularly in emotional expression and moral reasoning. However, challenges in accurate stance detection highlight the complexities of political communication modeling. Our novel analytical framework, including the Partisan Bias Divergence Assessment and the Partisan Class Tendency Prediction tasks, offers a comprehensive approach to evaluating the model's alignment with real-world political ideologies.

Looking forward, refining stance detection capabilities remains a priority, potentially through integrating more diverse datasets and enhancing contextual understanding. Future research could also explore adapting the model for cross-cultural and multilingual contexts, addressing ethical concerns and inherent biases, and expanding its application to real-time political discourse analysis. These directions not only promise to strengthen the model's utility in political discourse analysis but also contribute to the broader endeavor of creating ethically responsible and socially aware AI tools.

\bibliography{anthology,custom}
\bibliographystyle{acl_natbib}

\clearpage
\appendix

\section{Appendix} \label{app:appendices}

\begin{figure*}[!ht]
\centering

\begin{subfigure}{.44\textwidth}
  \centering
  \includegraphics[width=\linewidth]{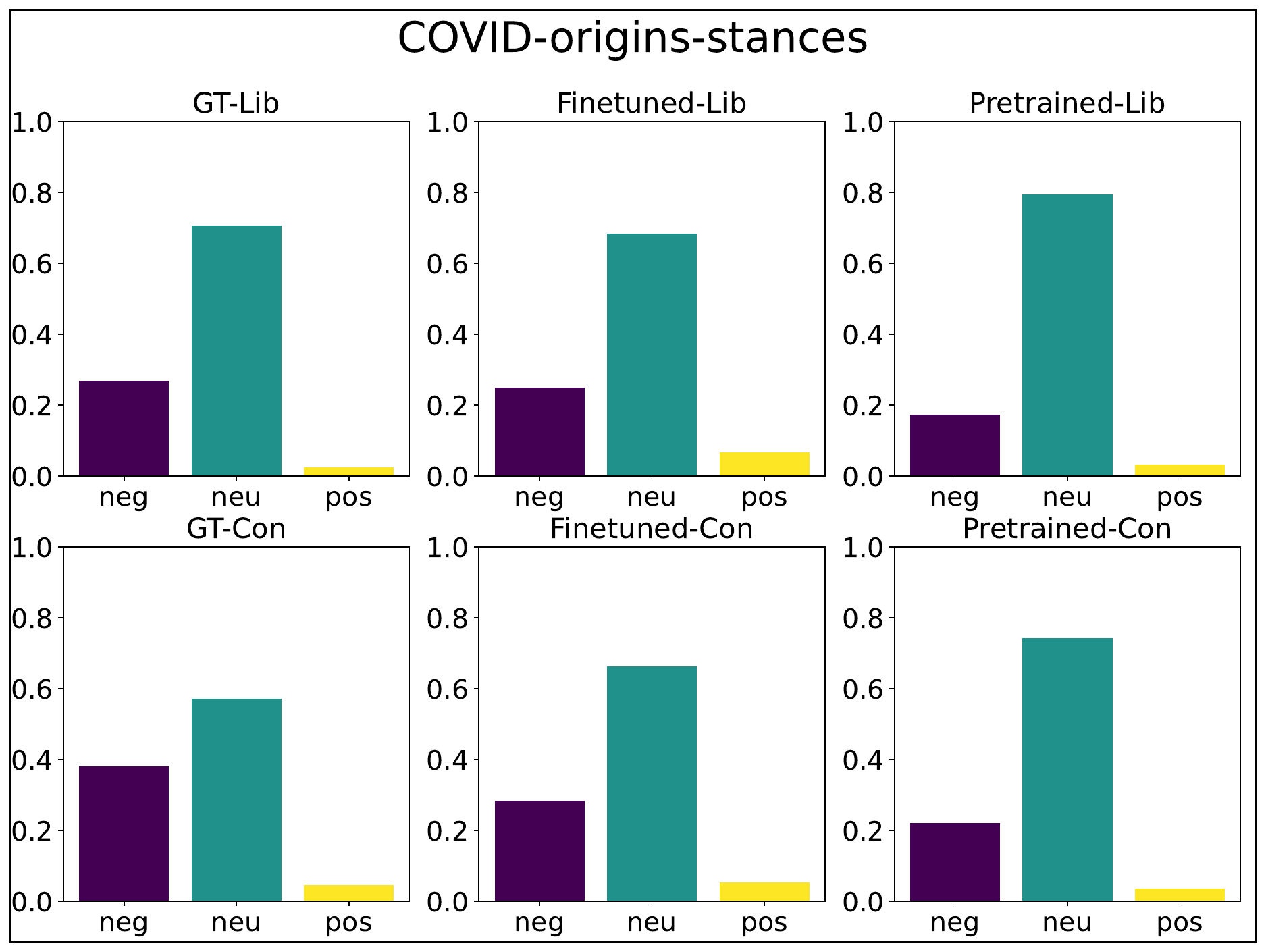}
\end{subfigure}%
\begin{subfigure}{.44\textwidth}
  \centering
  \includegraphics[width=\linewidth]{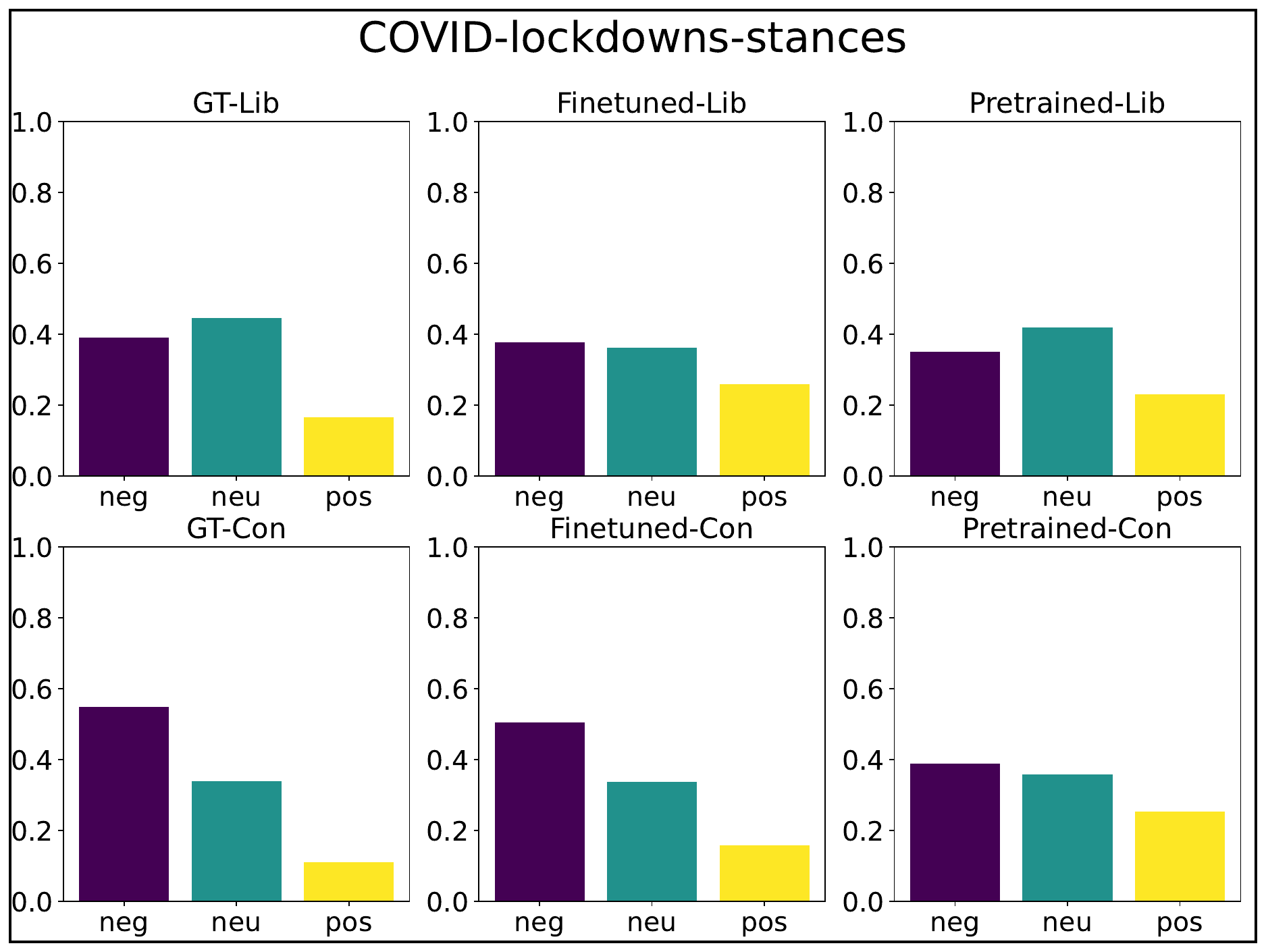}
\end{subfigure}%
\\

\begin{subfigure}{.44\textwidth}
  \centering
  \includegraphics[width=\linewidth]{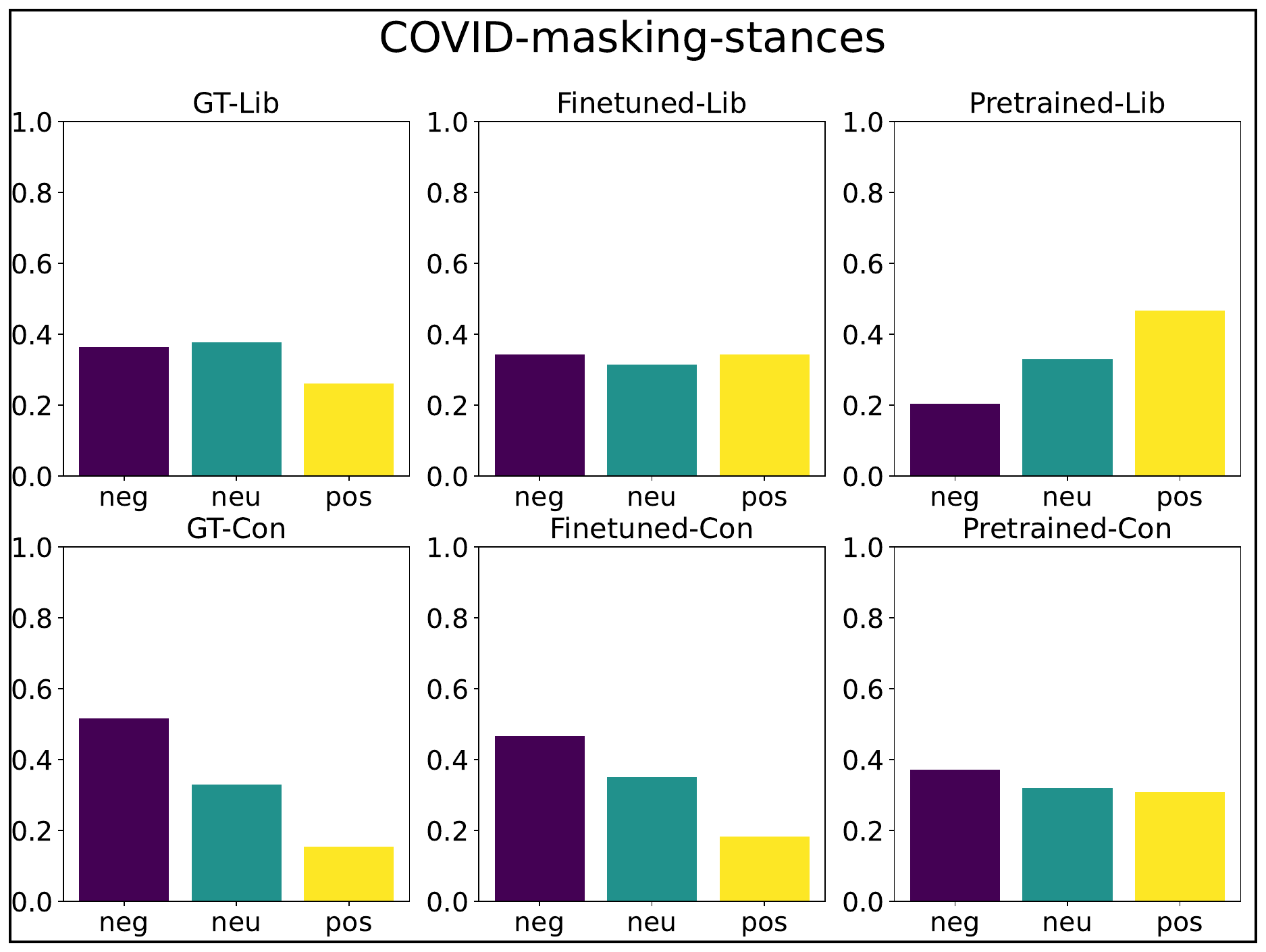}
\end{subfigure}%
\begin{subfigure}{.44\textwidth}
  \centering
  \includegraphics[width=\linewidth]{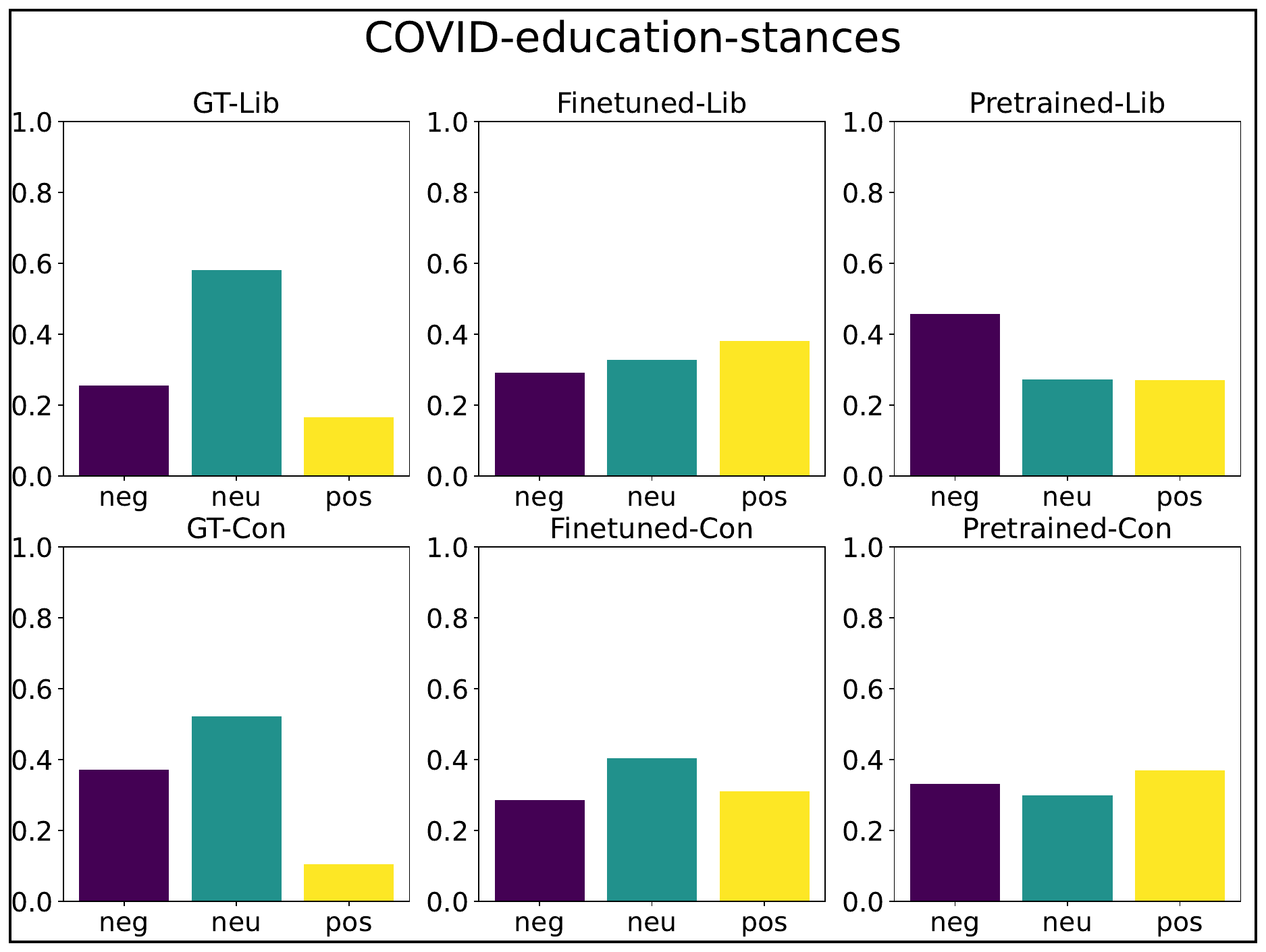}
\end{subfigure}
\\

\begin{subfigure}[!t]{.44\textwidth}
  \centering
  \includegraphics[width=\linewidth]{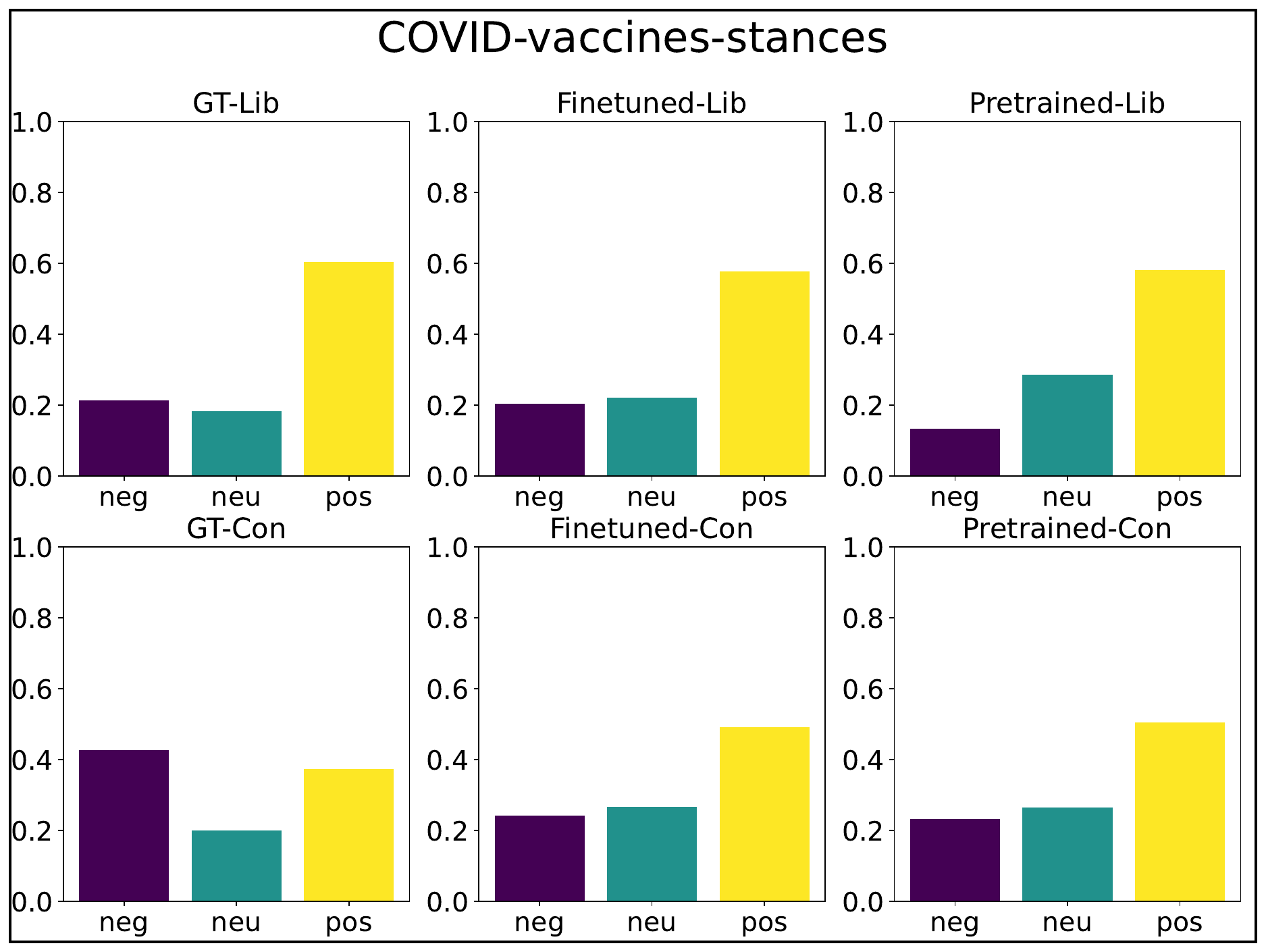}
\end{subfigure}%

\caption{Probability distribution of classes for stance detection in COVID-19 tweets.}
\label{fig:prob-stances-covid}

\end{figure*}

\begin{figure*}[ht]
\centering

\begin{subfigure}{.44\textwidth}
  \centering
  \includegraphics[width=\linewidth]{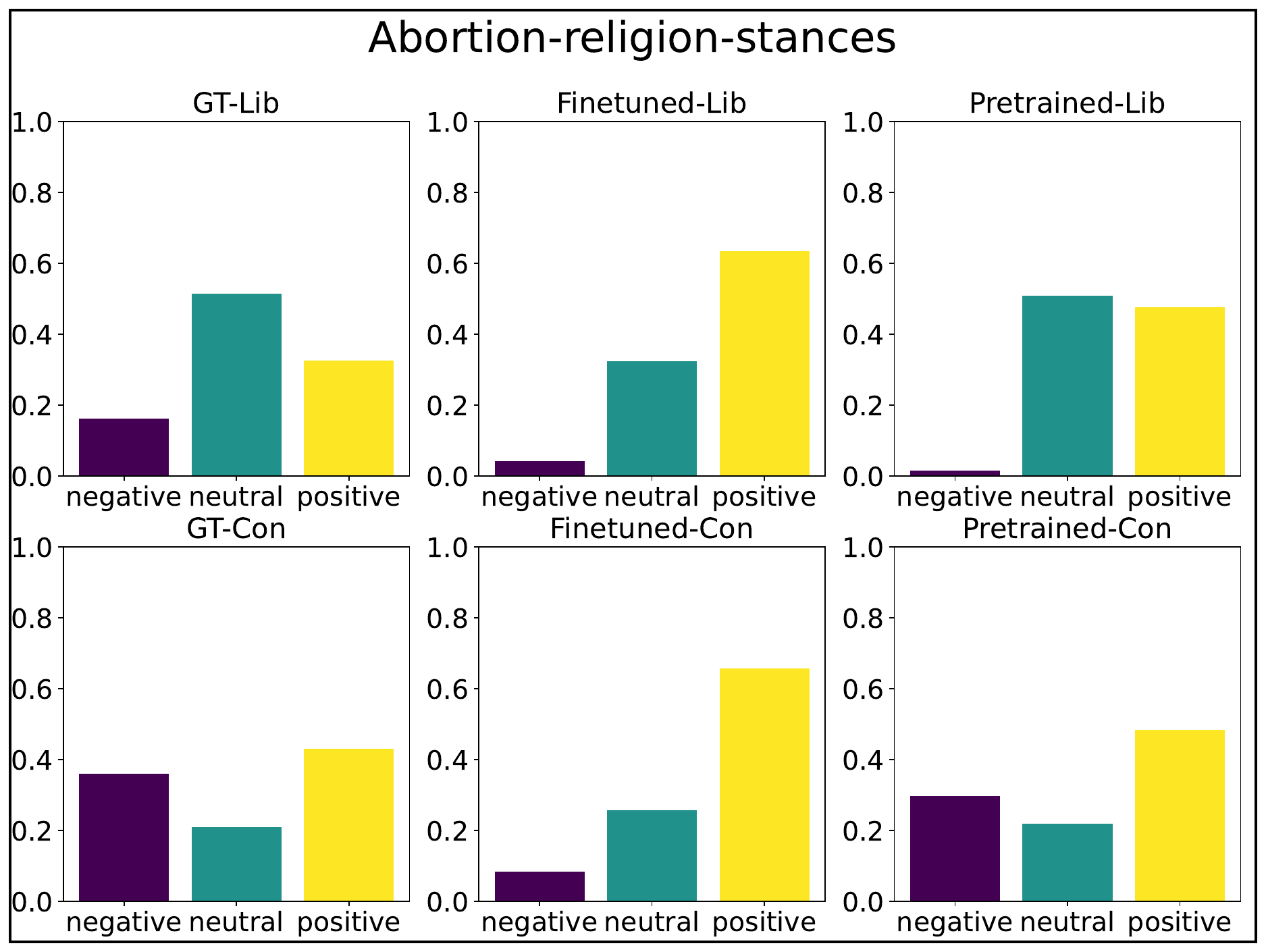}
\end{subfigure}%
\begin{subfigure}{.44\textwidth}
  \centering
  \includegraphics[width=\linewidth]{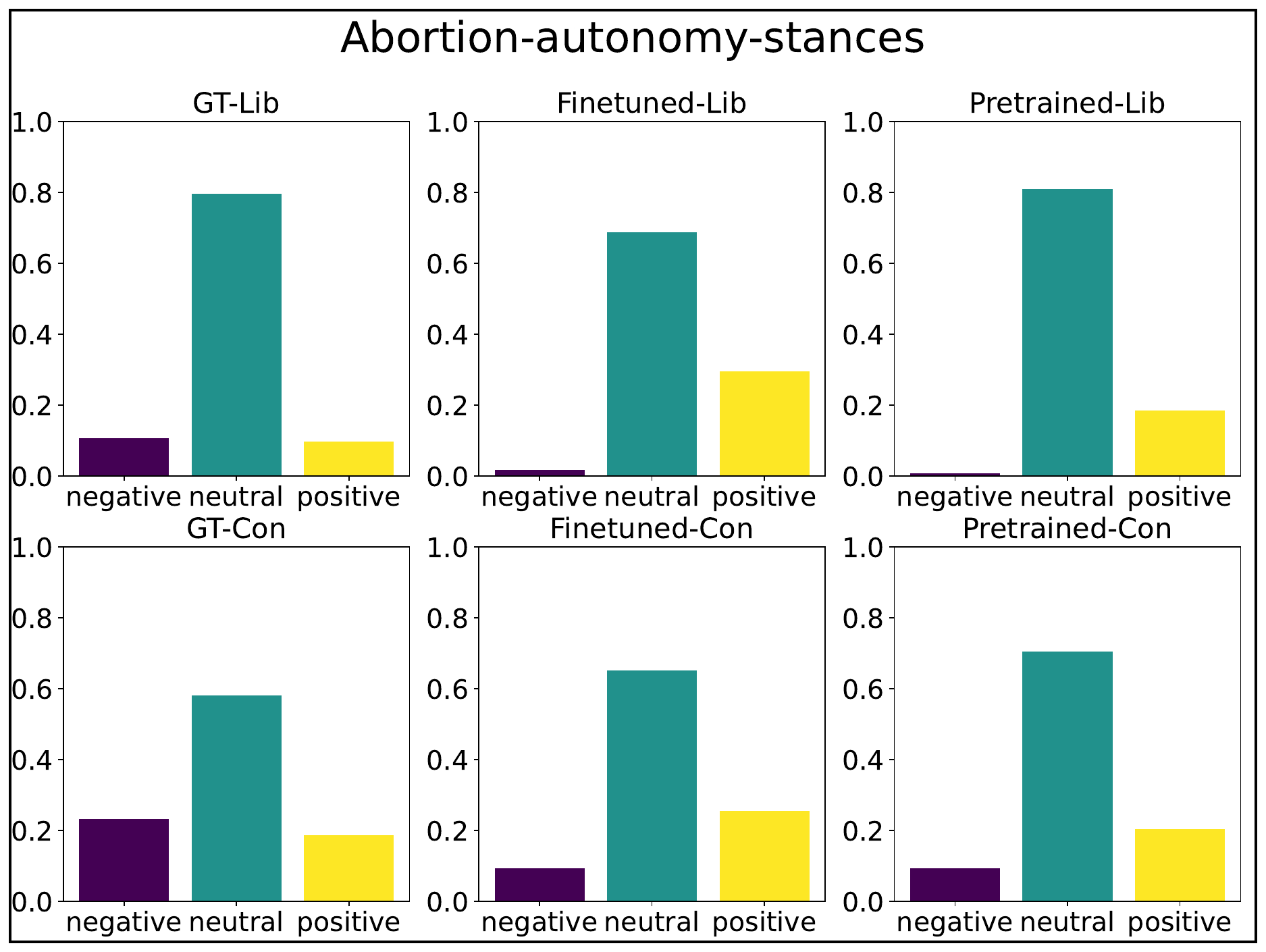}
\end{subfigure}%
\\

\begin{subfigure}{.44\textwidth}
  \centering
  \includegraphics[width=\linewidth]{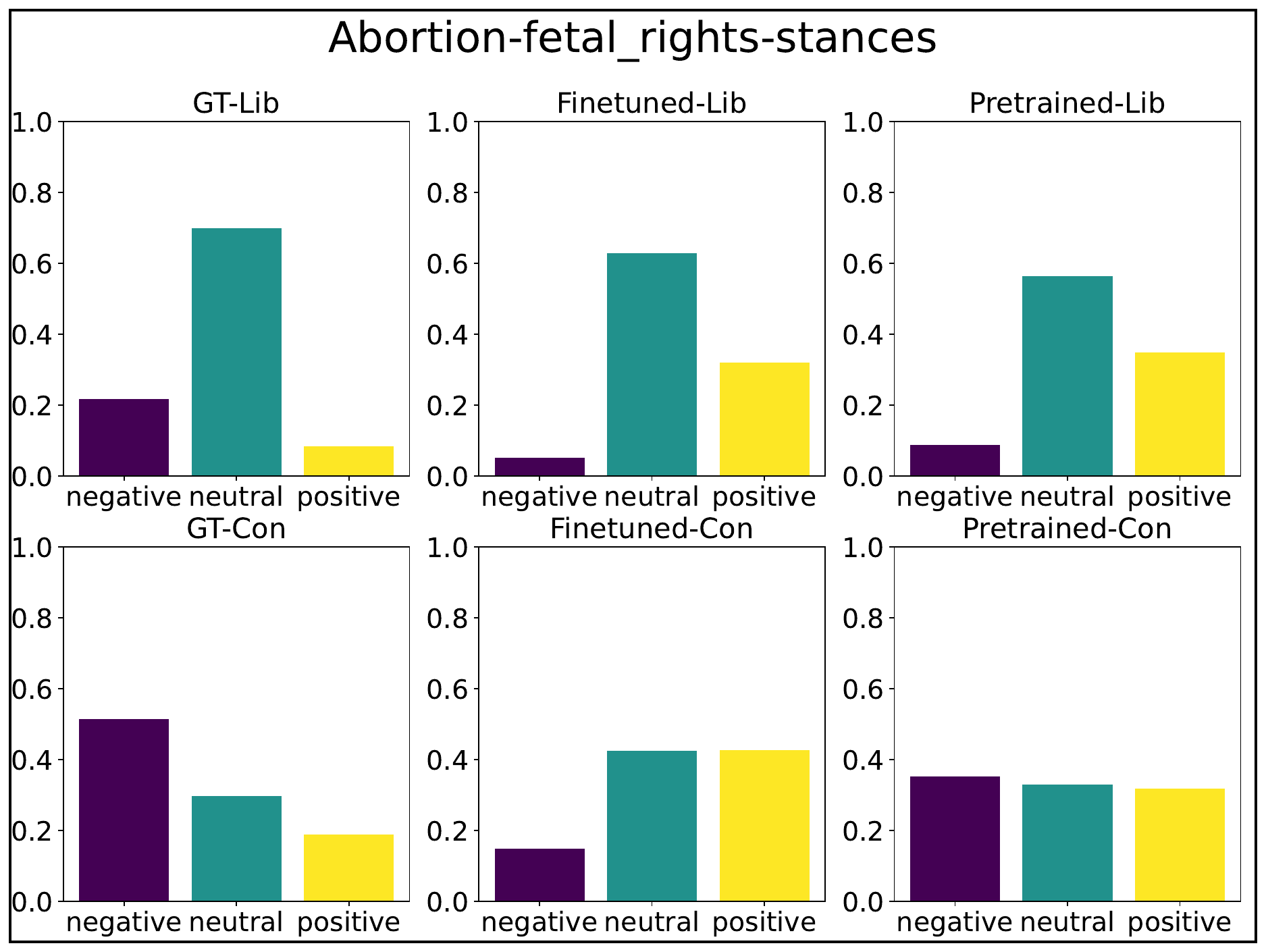}
\end{subfigure}%
\begin{subfigure}{.44\textwidth}
  \centering
  \includegraphics[width=\linewidth]{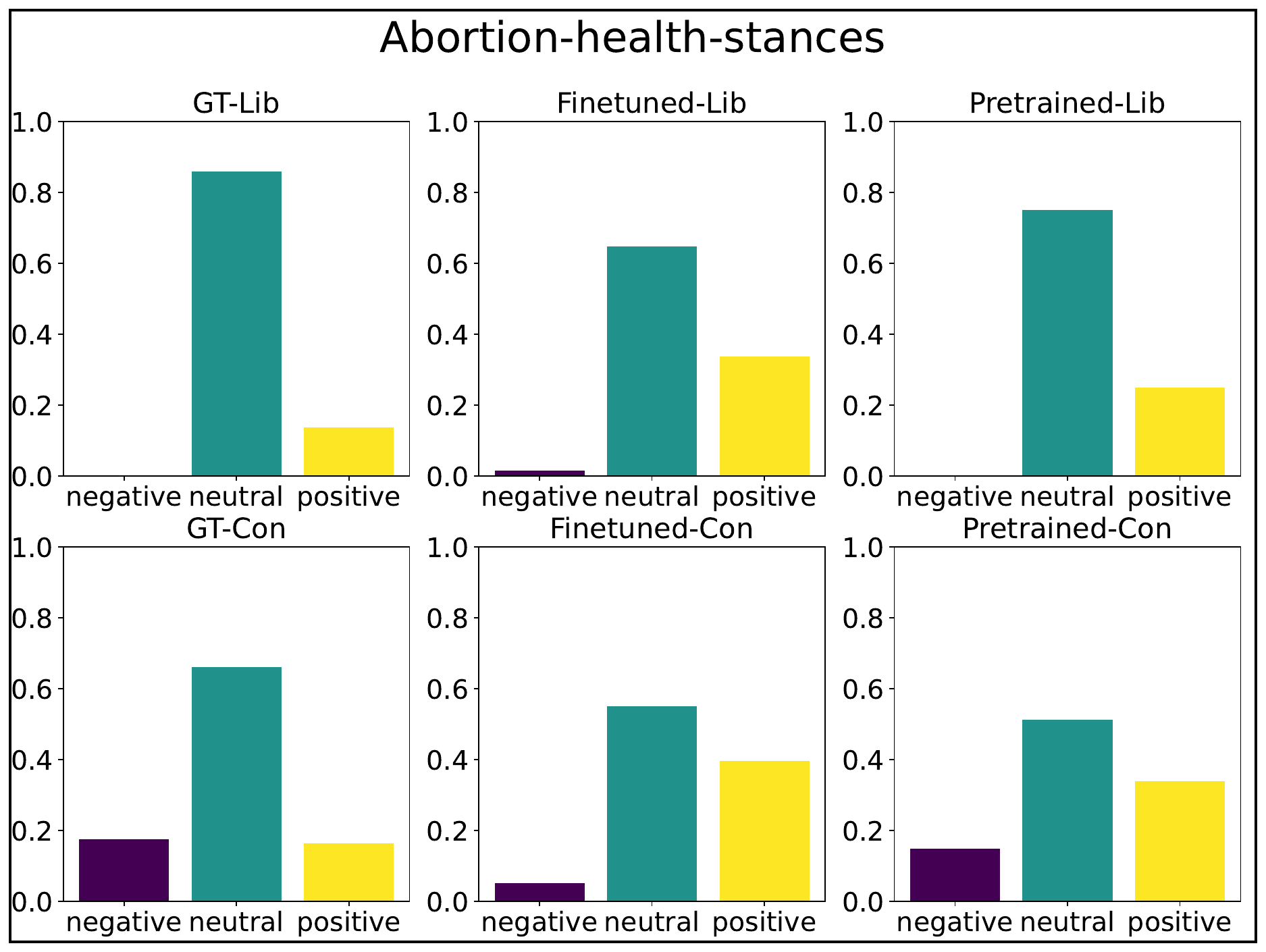}
\end{subfigure}
\\

\begin{subfigure}{.44\textwidth}
  \centering
  \includegraphics[width=\linewidth]{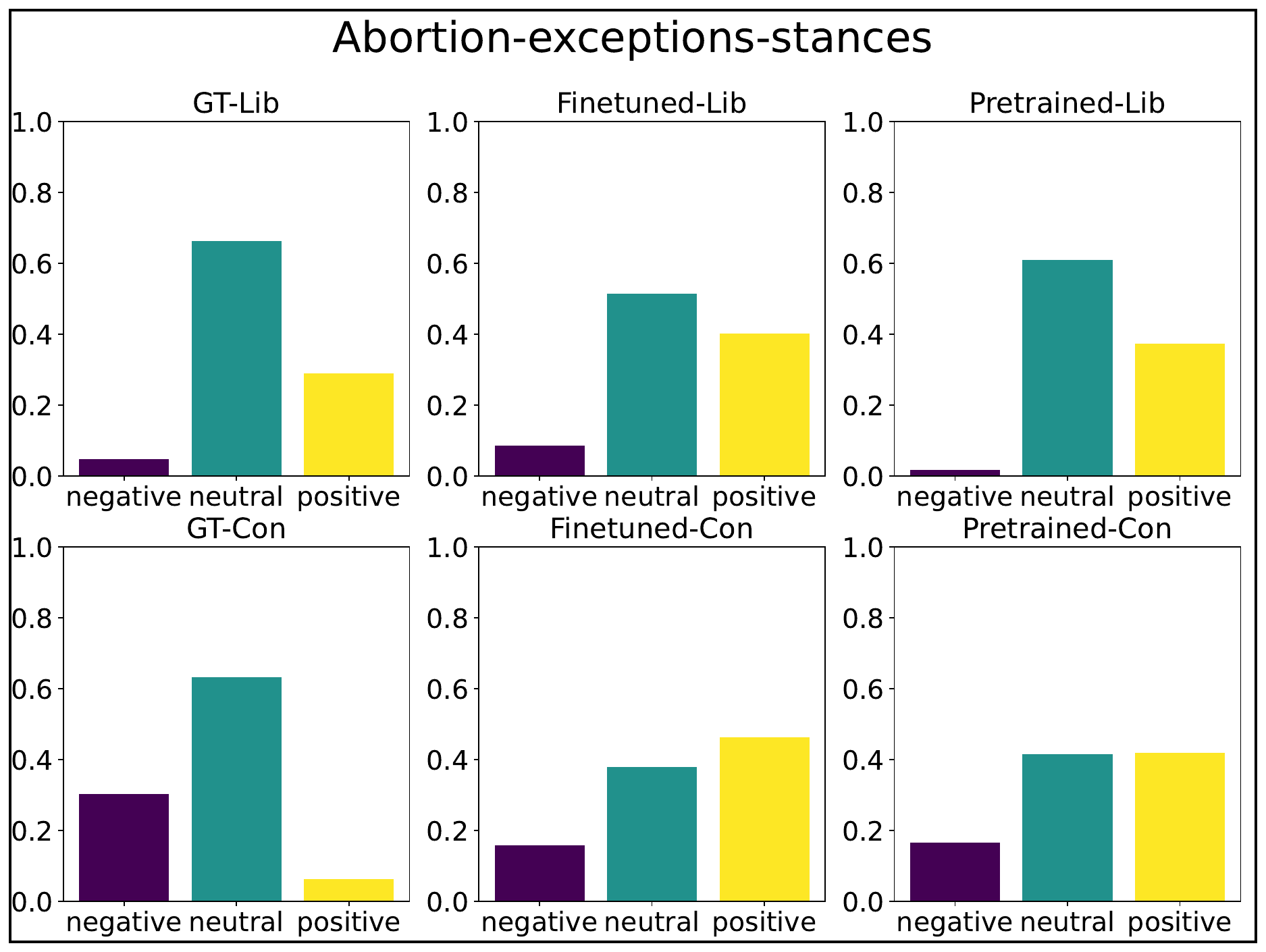}
\end{subfigure}%

\caption{Probability distribution of classes for stance detection in Abortion tweets.}
\label{fig:prob-stances-abortion}

\end{figure*}

\begin{figure*}[ht]
\centering

\begin{subfigure}{.44\textwidth}
  \centering
  \includegraphics[width=\linewidth]{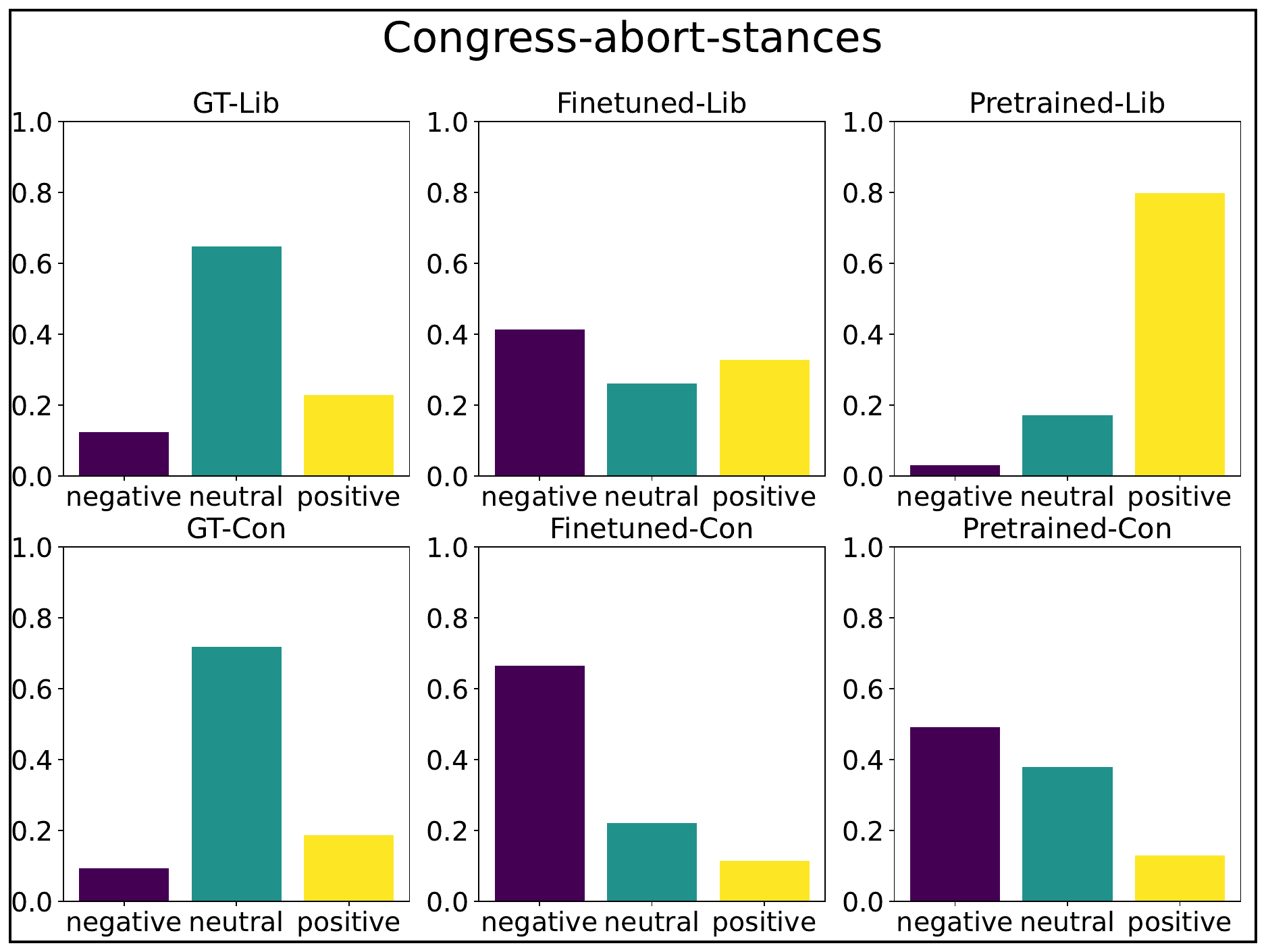}
\end{subfigure}%
\begin{subfigure}{.44\textwidth}
  \centering
  \includegraphics[width=\linewidth]{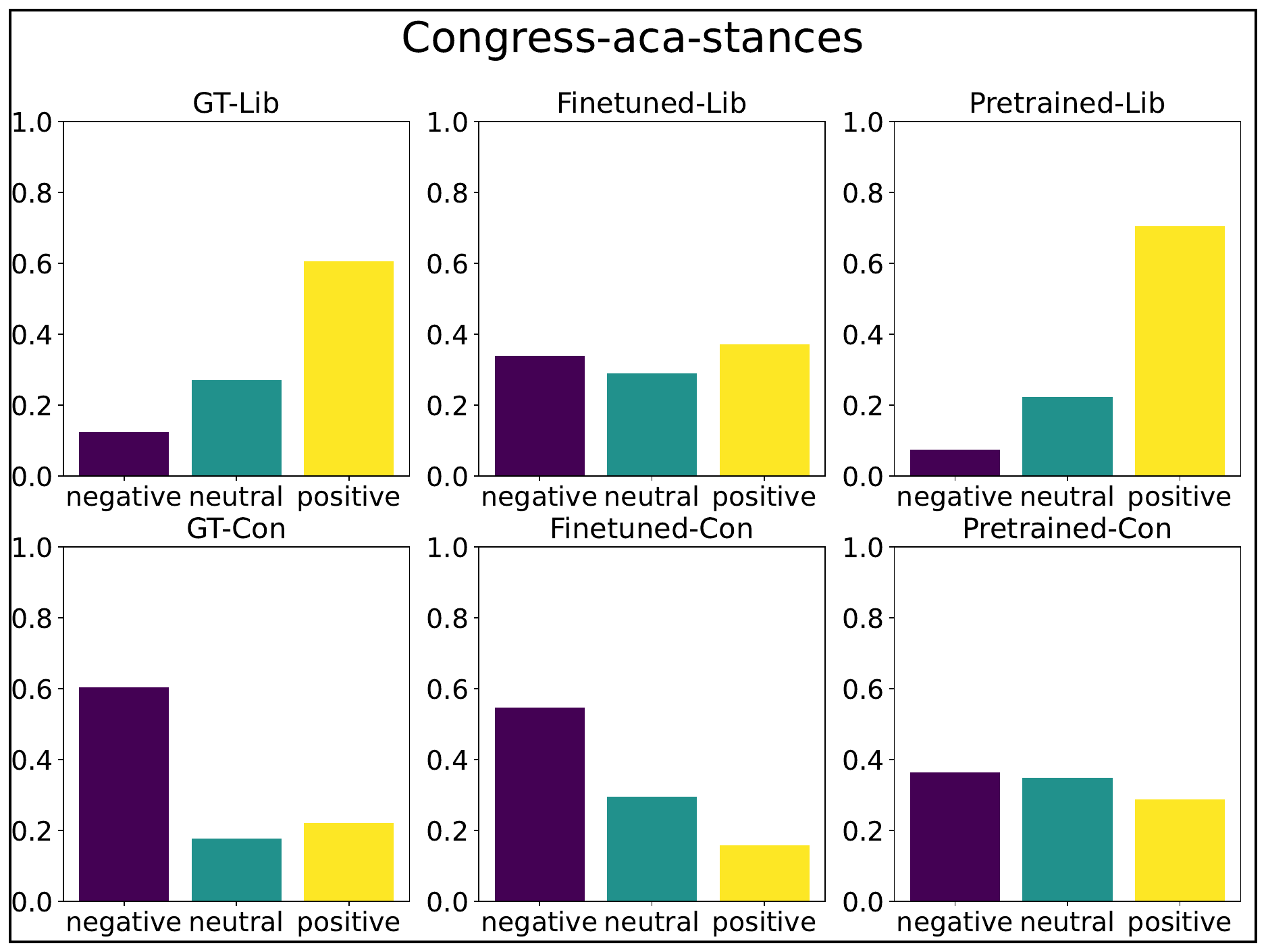}
\end{subfigure}%
\\

\begin{subfigure}{.44\textwidth}
  \centering
  \includegraphics[width=\linewidth]{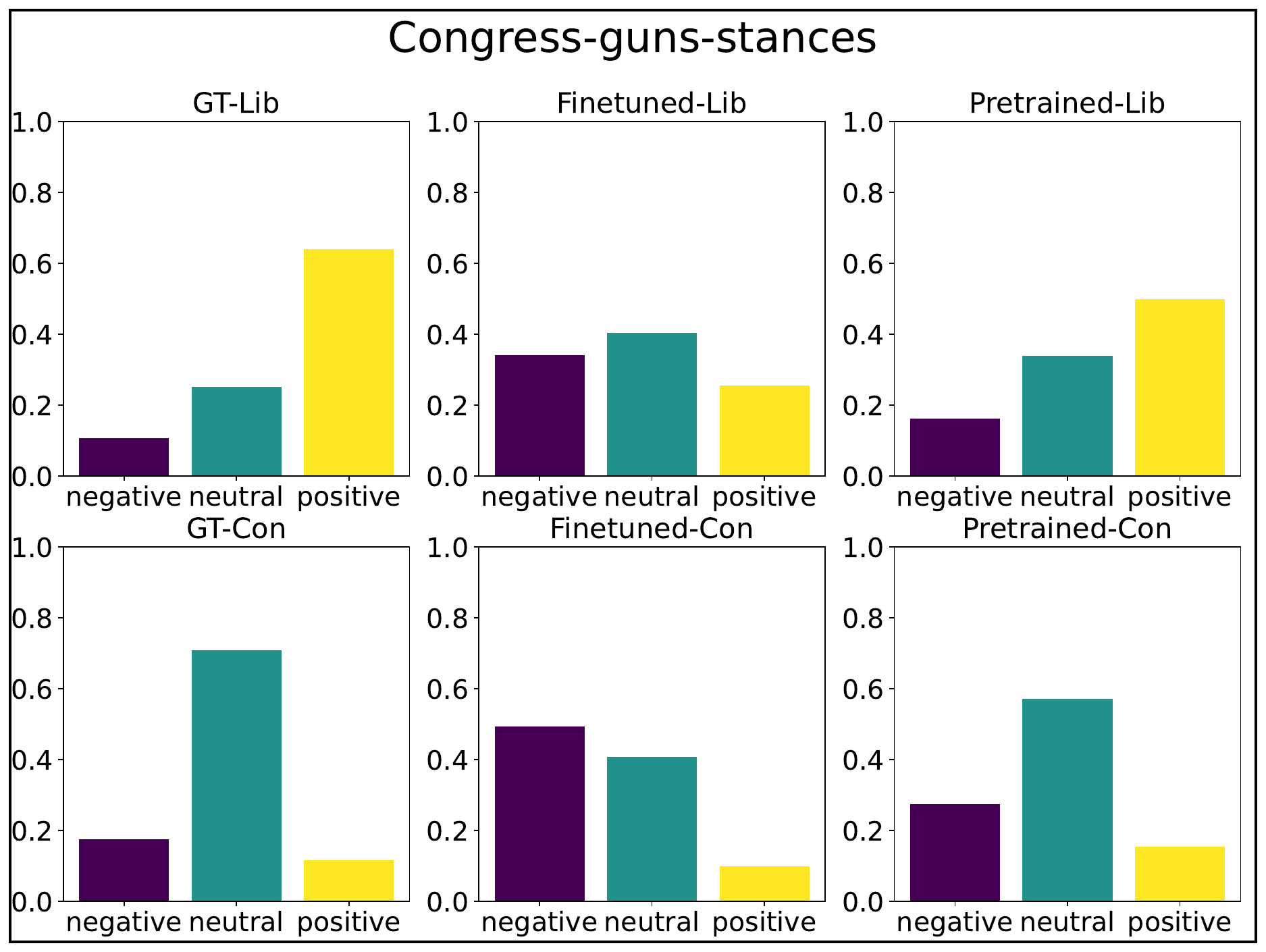}
\end{subfigure}%
\begin{subfigure}{.44\textwidth}
  \centering
  \includegraphics[width=\linewidth]{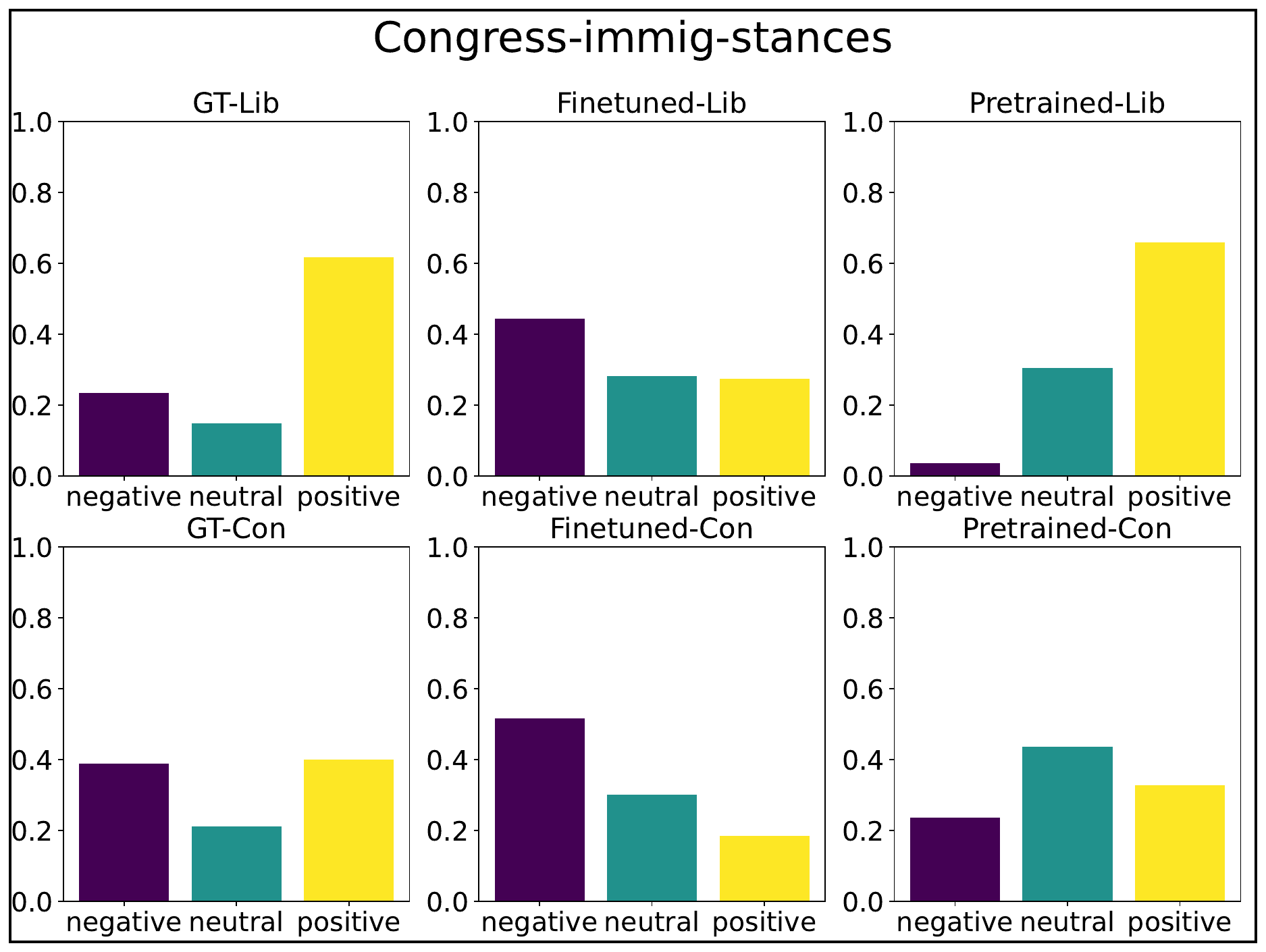}
\end{subfigure}
\\

\begin{subfigure}{.44\textwidth}
  \centering
  \includegraphics[width=\linewidth]{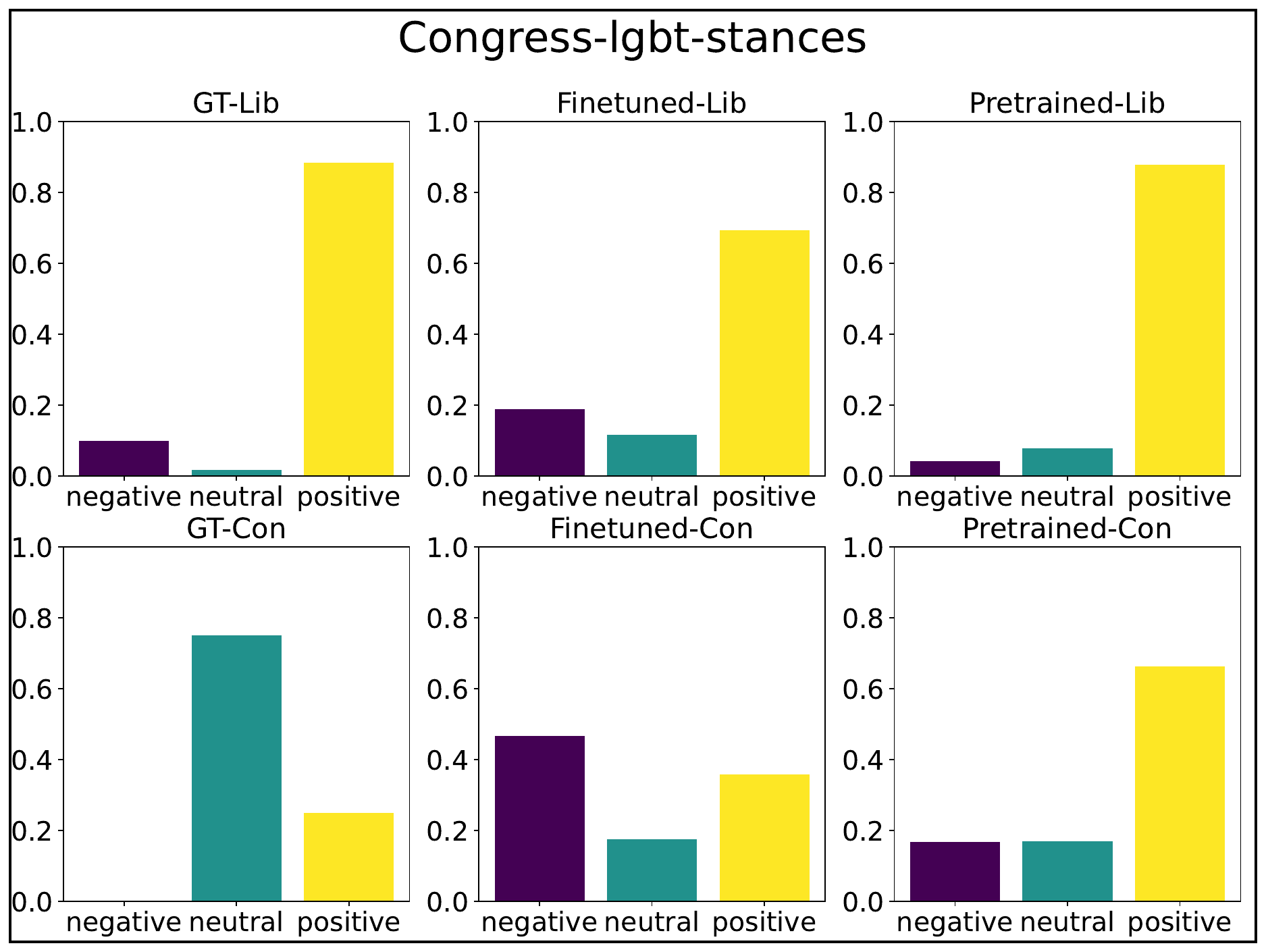}
\end{subfigure}%
\begin{subfigure}{.44\textwidth}
  \centering
  \includegraphics[width=\linewidth]{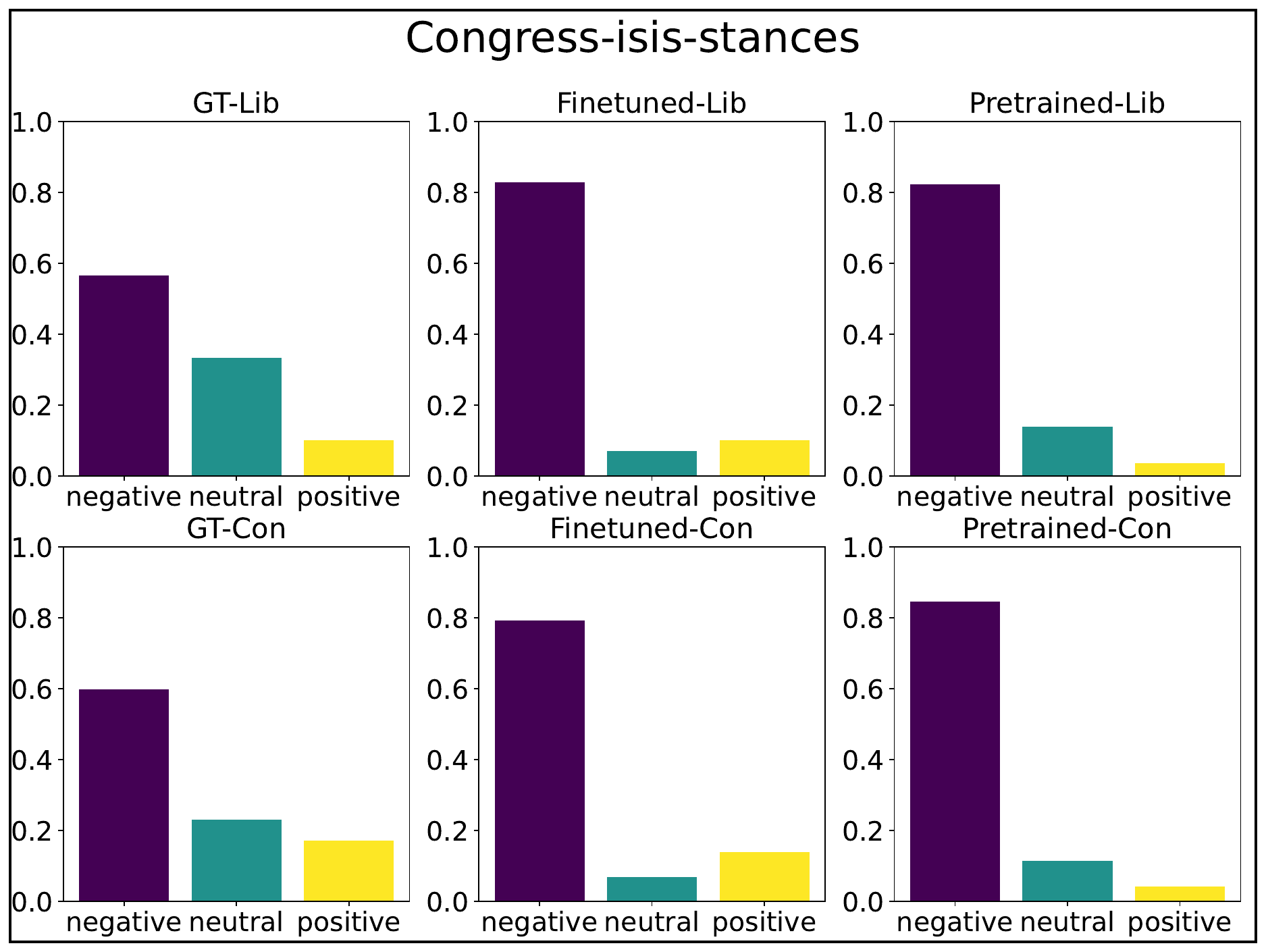}
\end{subfigure}%

\caption{Probability distribution of classes for stance detection in COVID-19 tweets.}
\label{fig:prob-stances-congress}

\end{figure*}


\begin{figure*}[ht]
\centering

\begin{subfigure}{.44\textwidth}
  \centering
  \includegraphics[width=\linewidth]{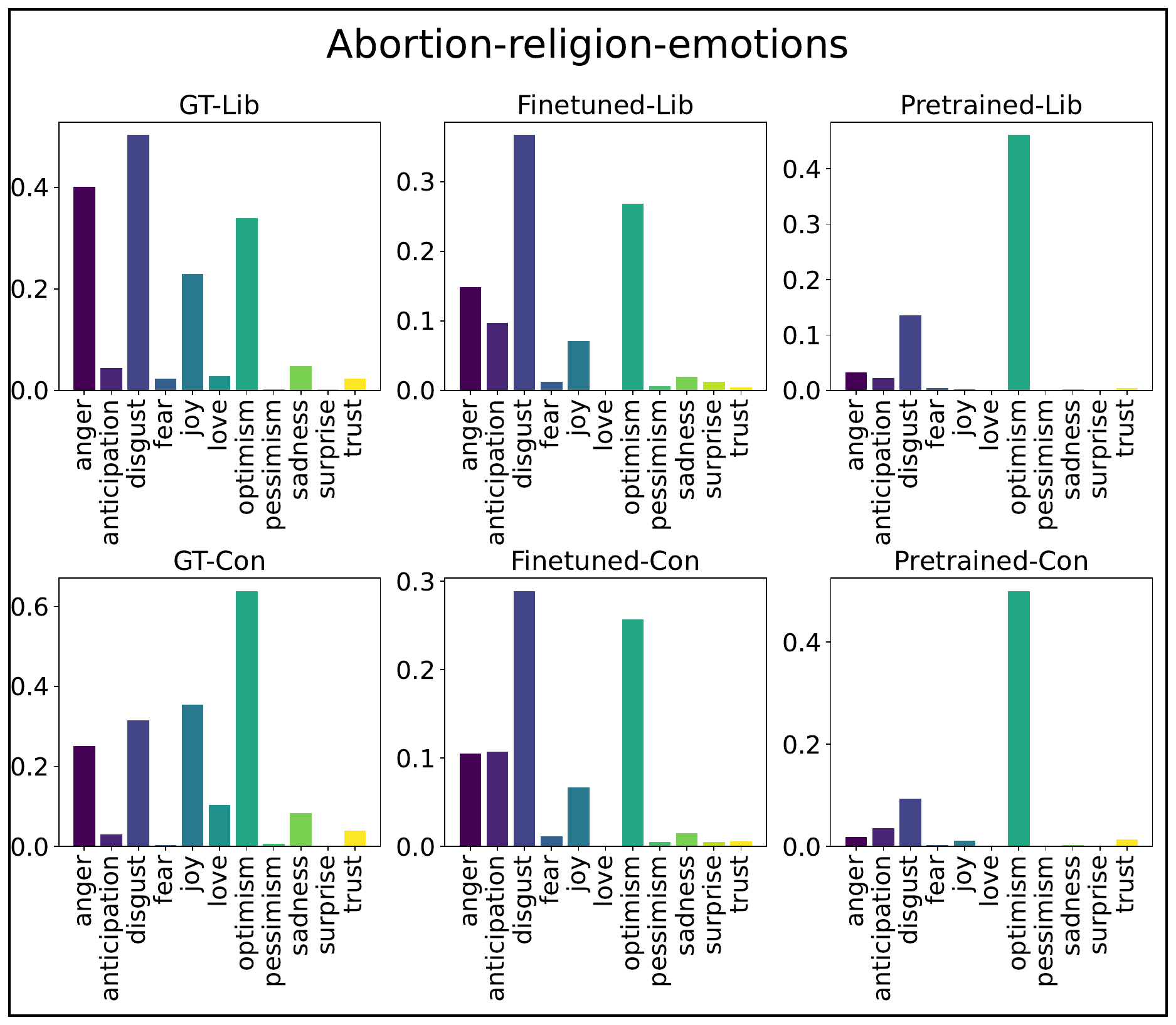}
\end{subfigure}%
\begin{subfigure}{.44\textwidth}
  \centering
  \includegraphics[width=\linewidth]{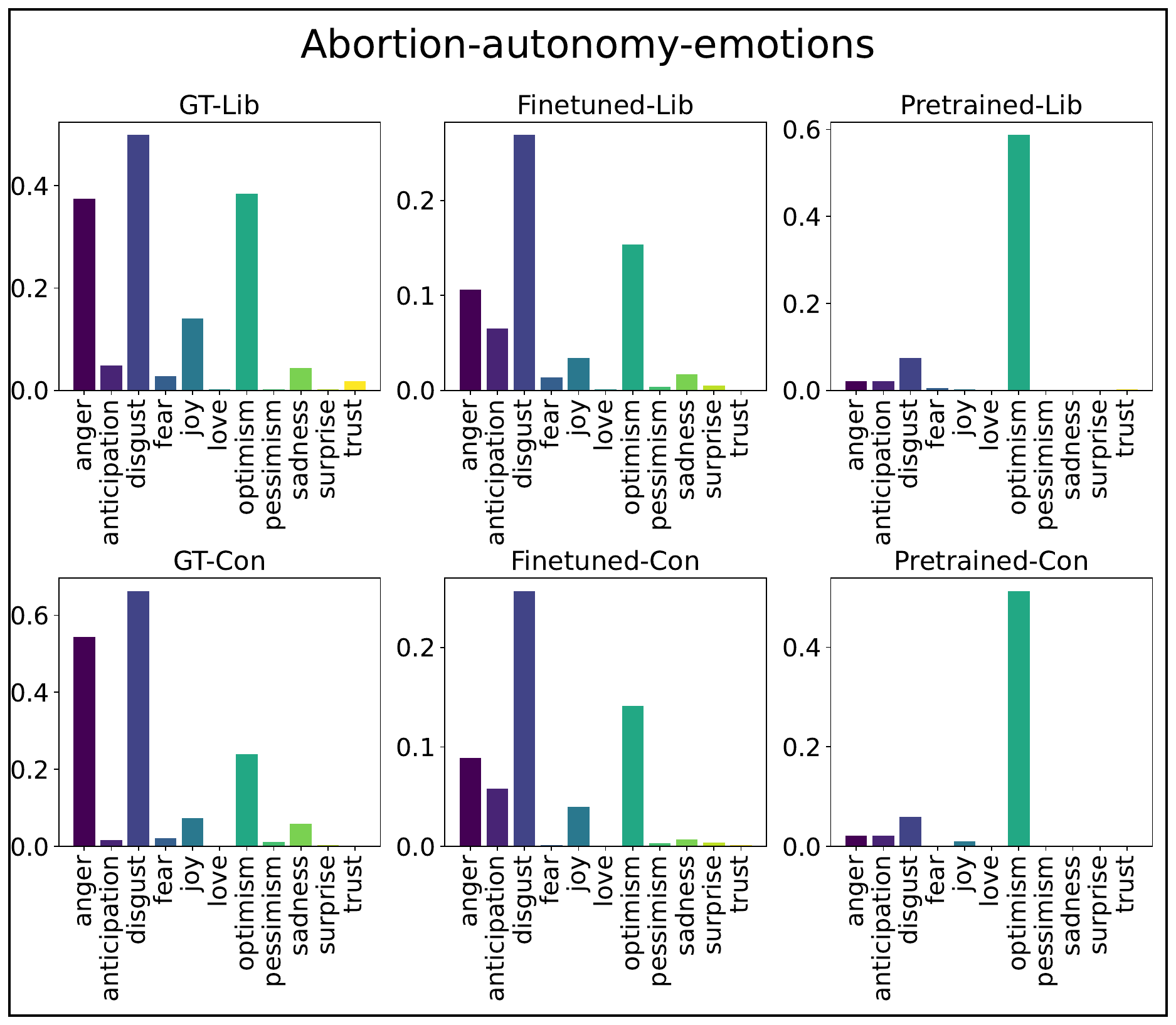}
\end{subfigure}%
\\

\begin{subfigure}{.44\textwidth}
  \centering
  \includegraphics[width=\linewidth]{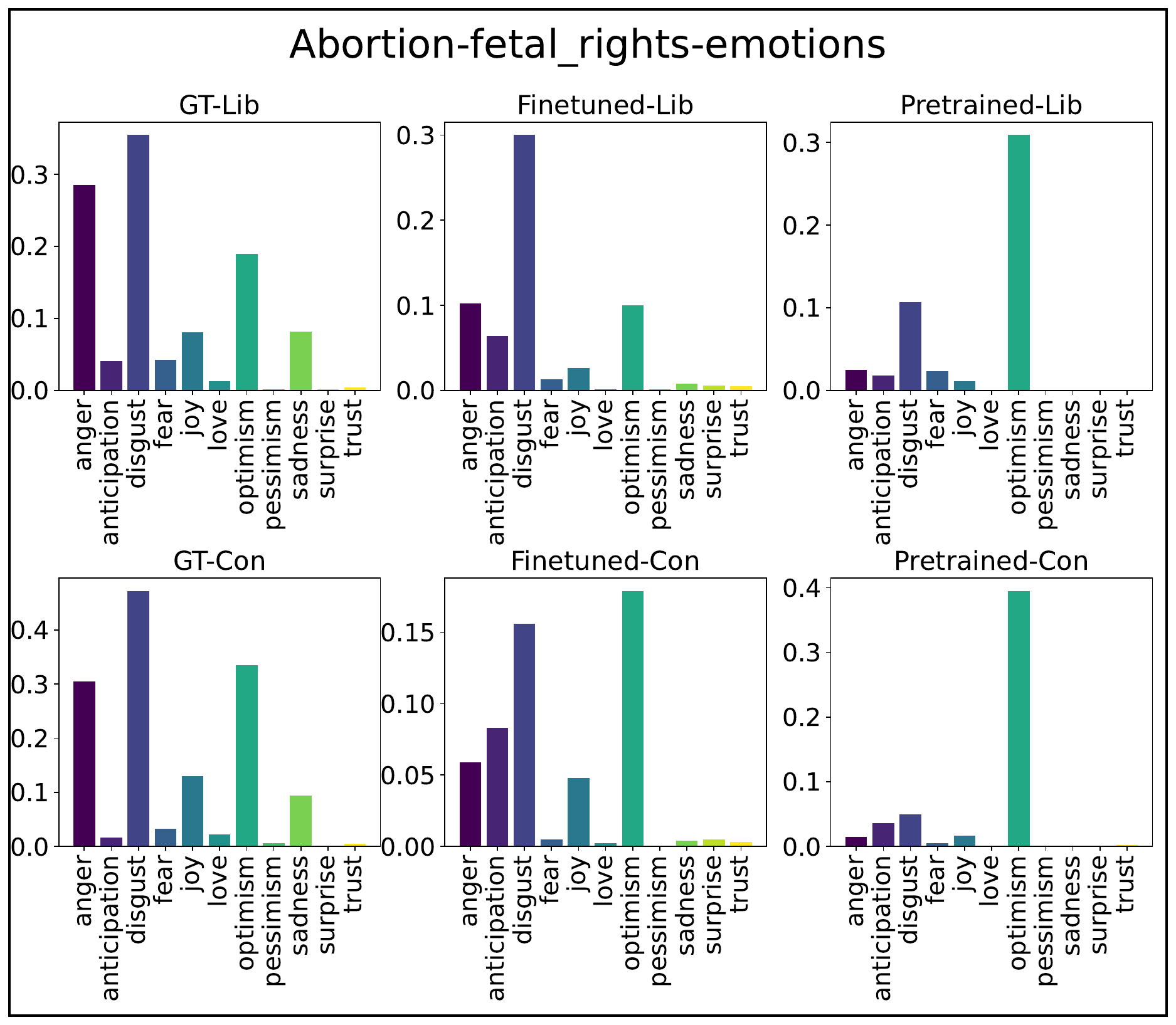}
\end{subfigure}%
\begin{subfigure}{.44\textwidth}
  \centering
  \includegraphics[width=\linewidth]{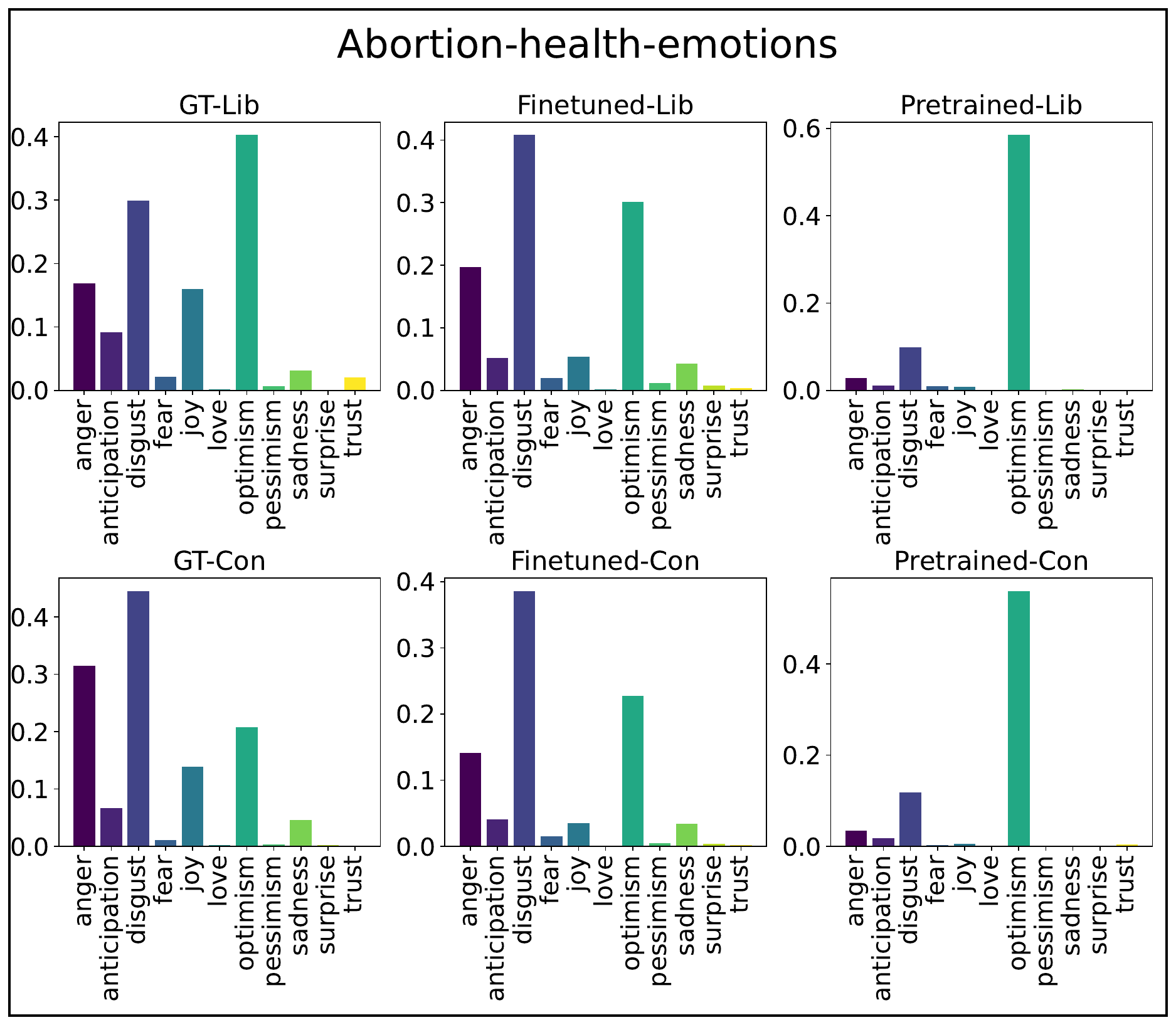}
\end{subfigure}
\\

\begin{subfigure}{.44\textwidth}
  \centering
  \includegraphics[width=\linewidth]{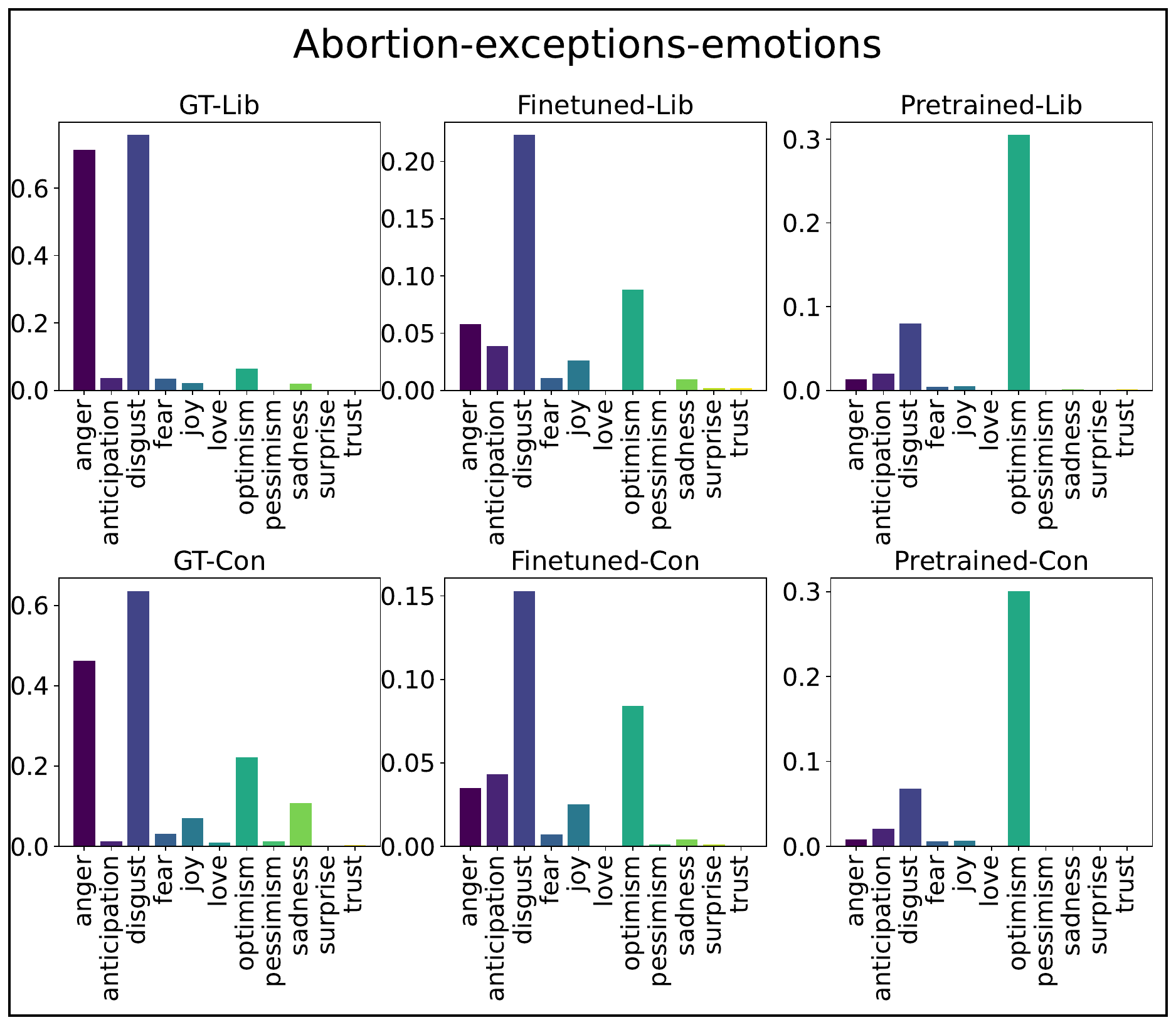}
\end{subfigure}%

\caption{Probability distribution of classes for emotion detection in Abortion tweets.}
\label{fig:prob-emotions-abortion}

\end{figure*}

\begin{figure*}[ht]
\centering

\begin{subfigure}{.44\textwidth}
  \centering
  \includegraphics[width=\linewidth]{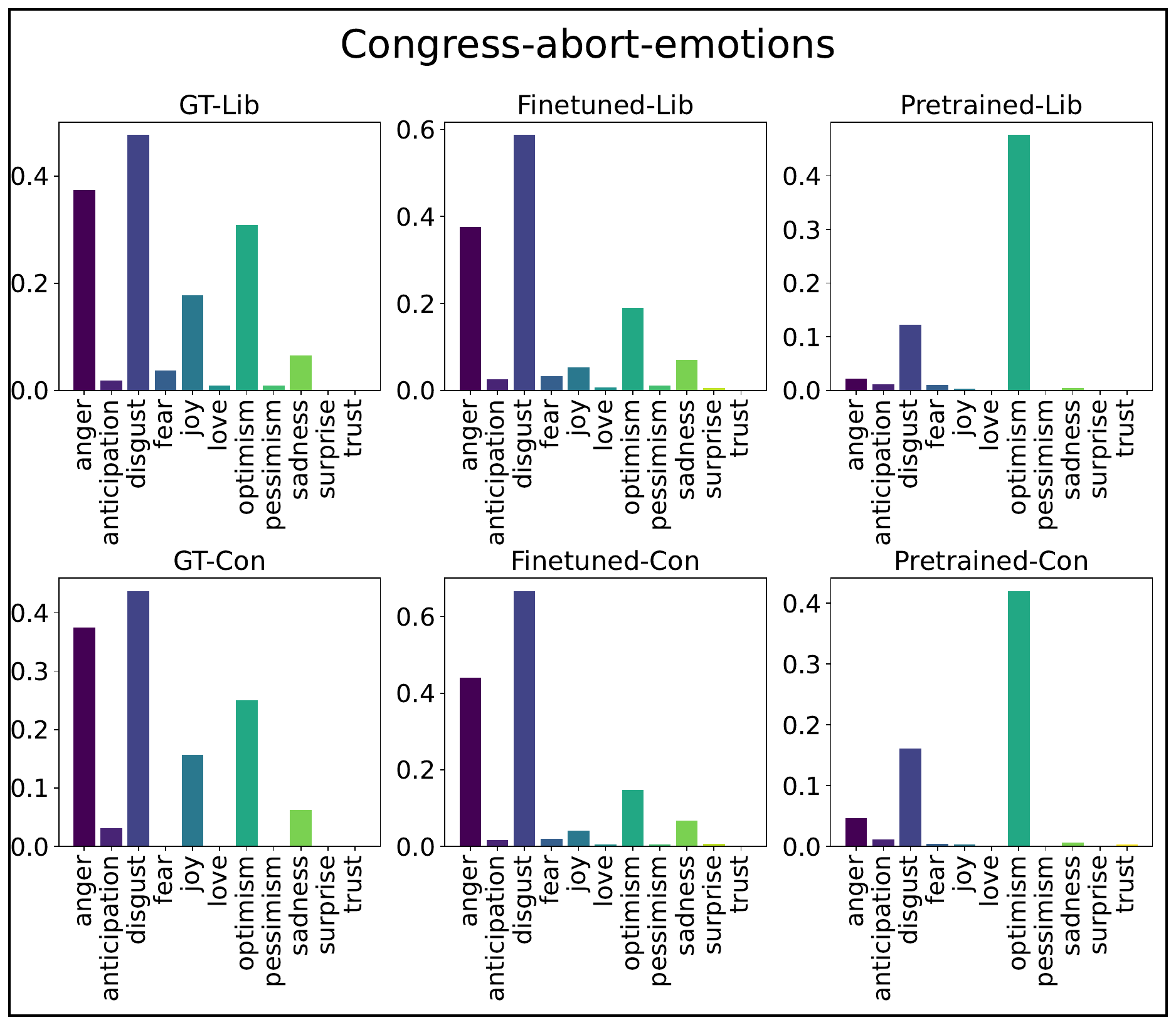}
\end{subfigure}%
\begin{subfigure}{.44\textwidth}
  \centering
  \includegraphics[width=\linewidth]{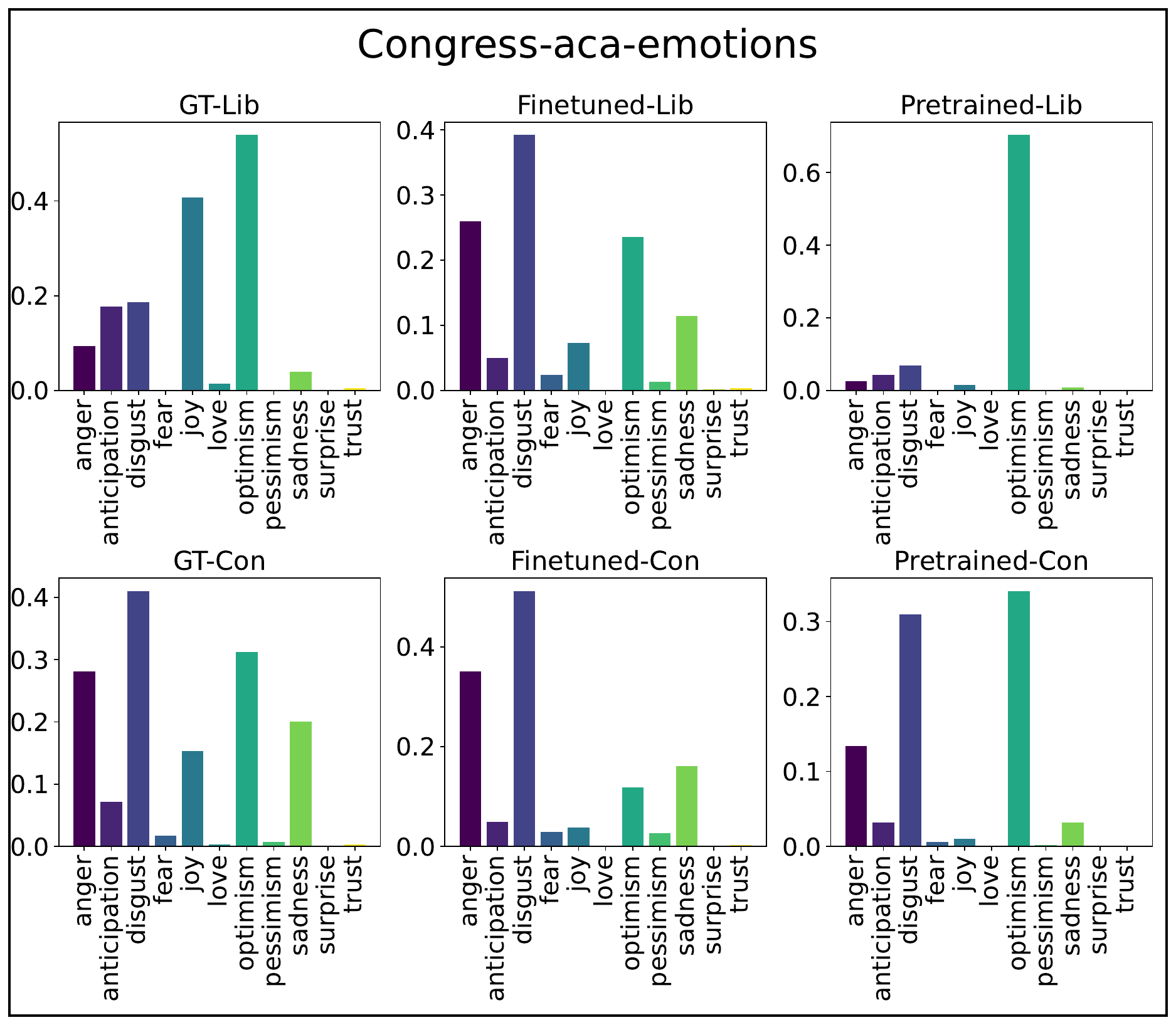}
\end{subfigure}%
\\

\begin{subfigure}{.44\textwidth}
  \centering
  \includegraphics[width=\linewidth]{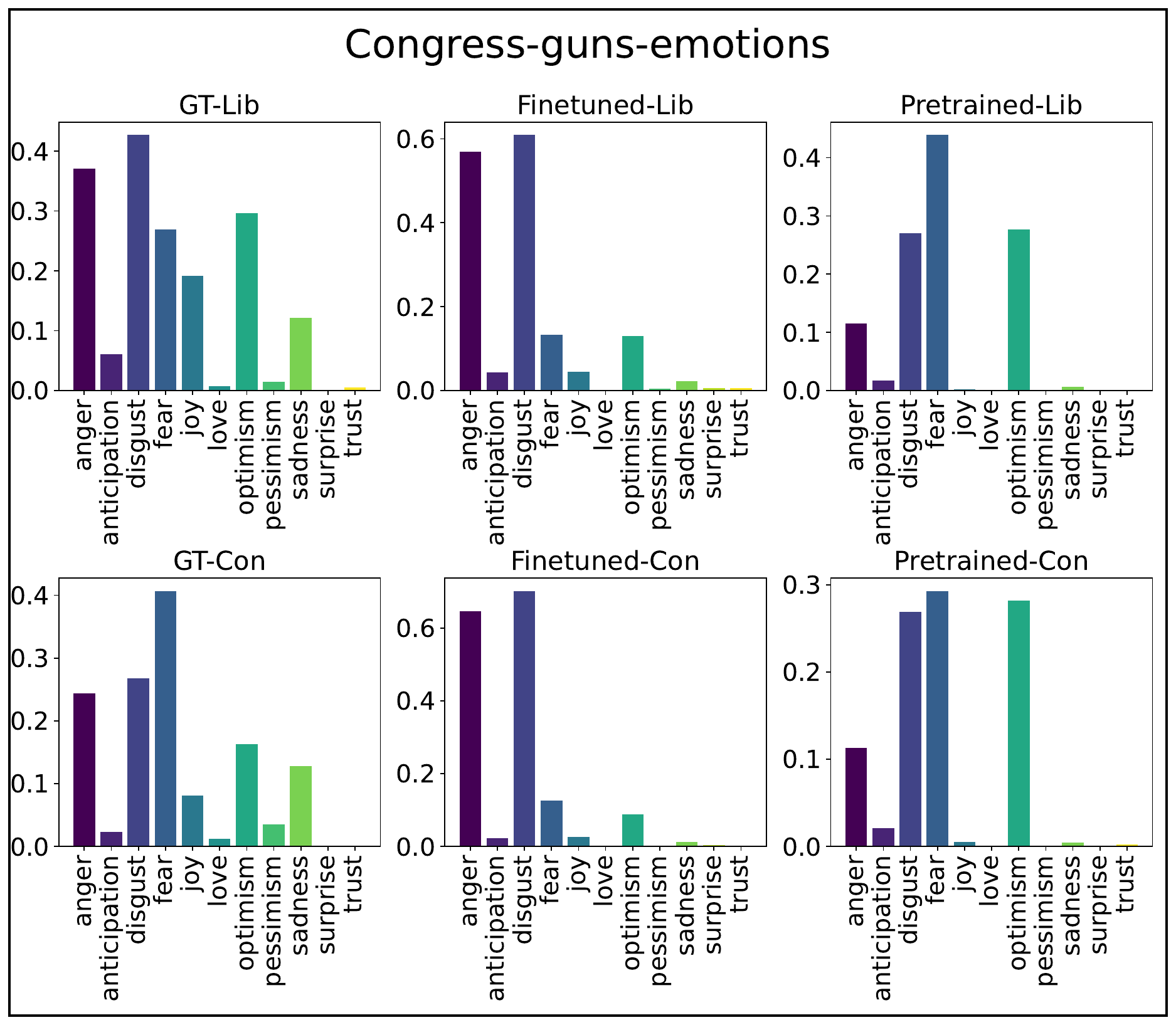}
\end{subfigure}%
\begin{subfigure}{.44\textwidth}
  \centering
  \includegraphics[width=\linewidth]{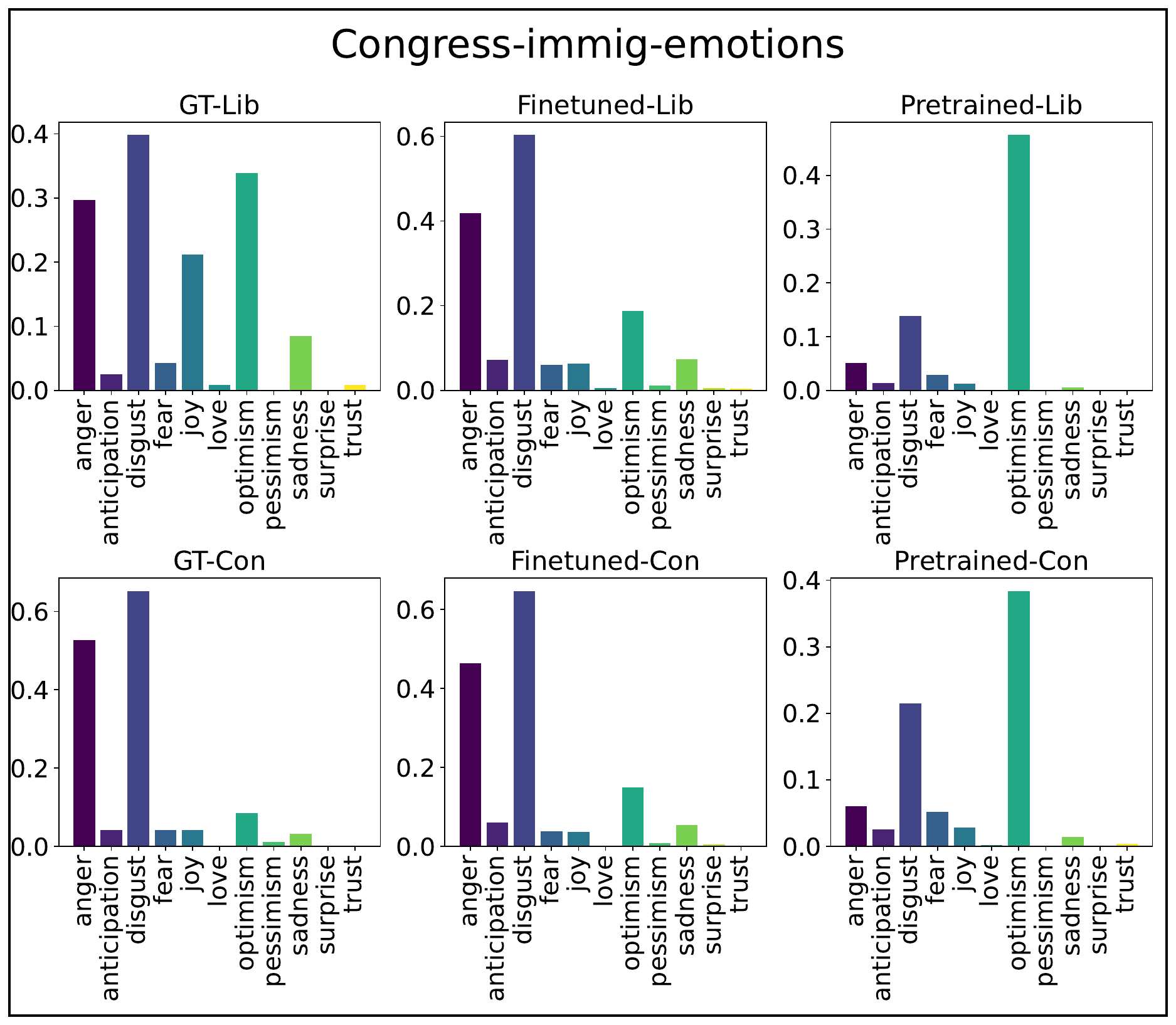}
\end{subfigure}
\\

\begin{subfigure}{.44\textwidth}
  \centering
  \includegraphics[width=\linewidth]{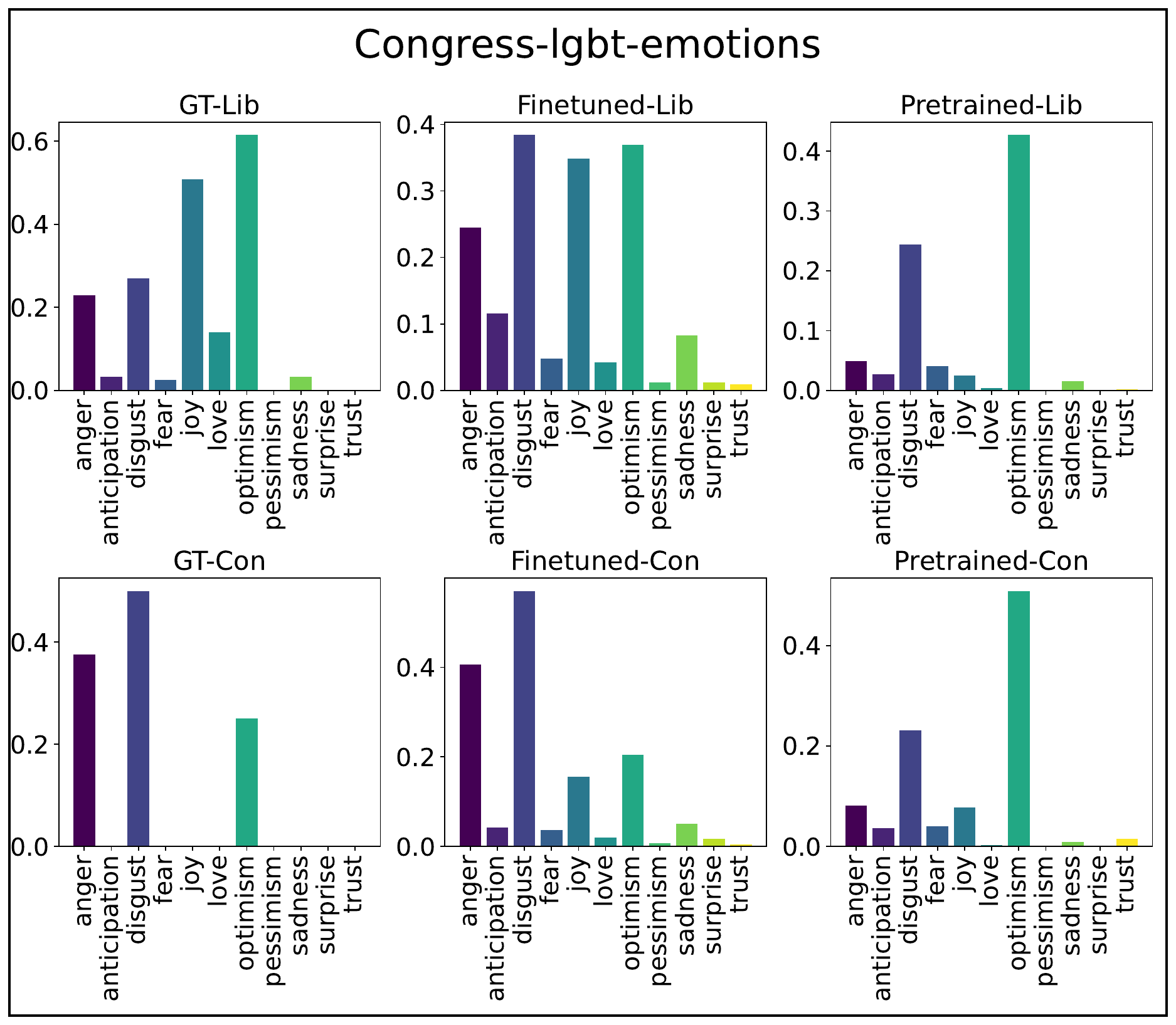}
\end{subfigure}%
\begin{subfigure}{.44\textwidth}
  \centering
  \includegraphics[width=\linewidth]{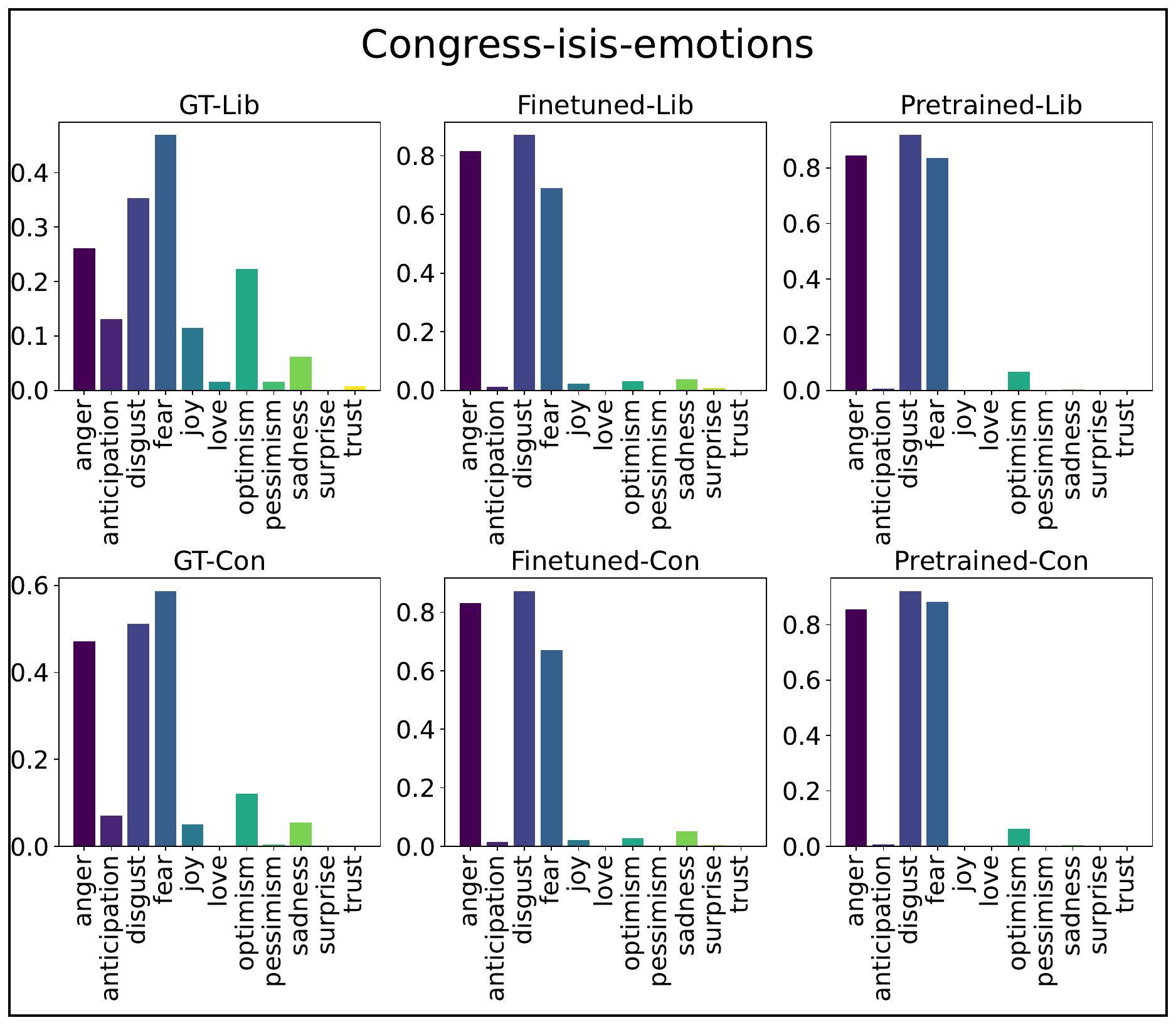}
\end{subfigure}%

\caption{Probability distribution of classes for emotion detection in congress tweets.}
\label{fig:prob-emotions-congress}

\end{figure*}


\begin{figure*}[ht]
\centering

\begin{subfigure}{.44\textwidth}
  \centering
  \includegraphics[width=\linewidth]{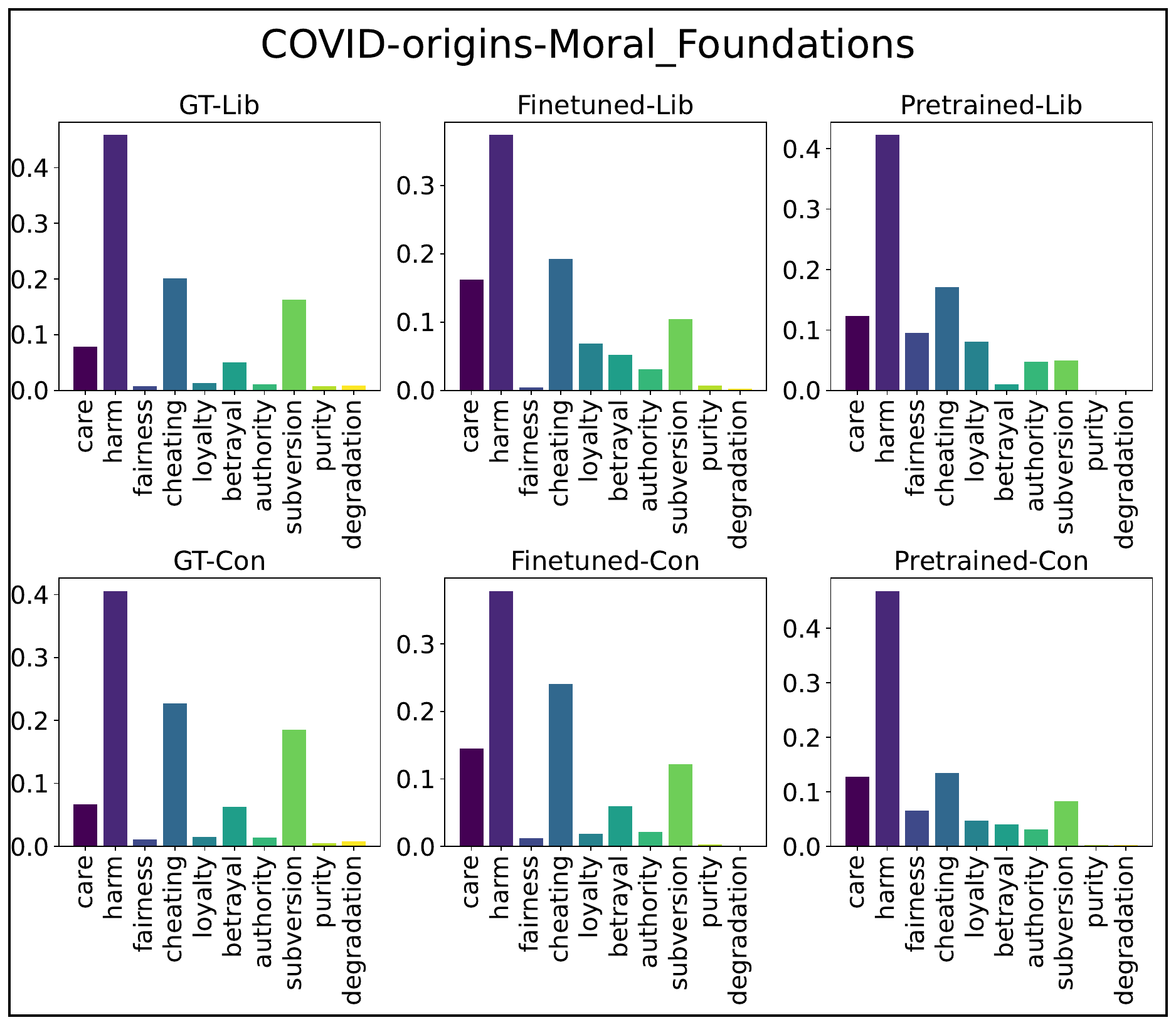}
\end{subfigure}%
\begin{subfigure}{.44\textwidth}
  \centering
  \includegraphics[width=\linewidth]{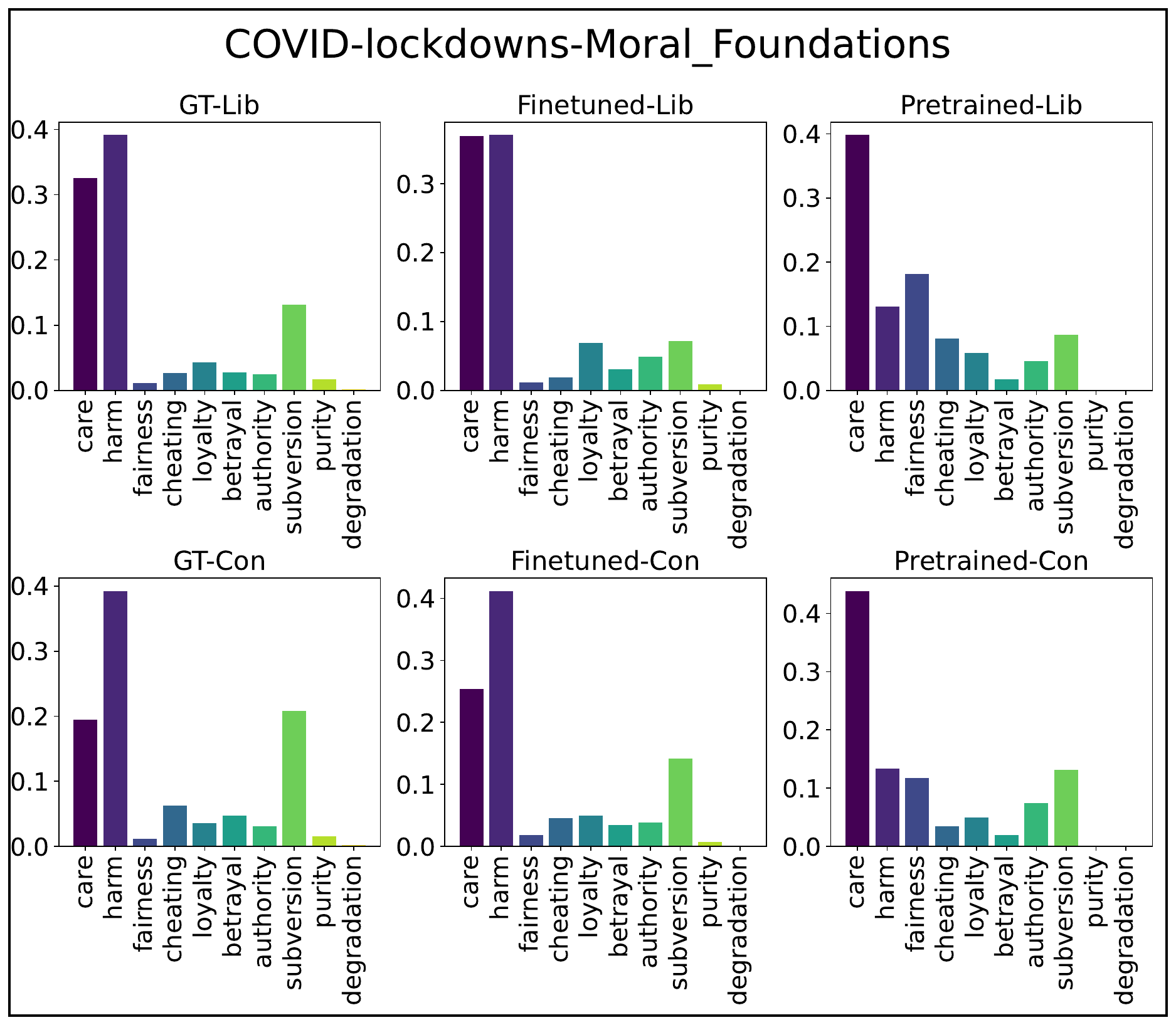}
\end{subfigure}%
\\

\begin{subfigure}{.44\textwidth}
  \centering
  \includegraphics[width=\linewidth]{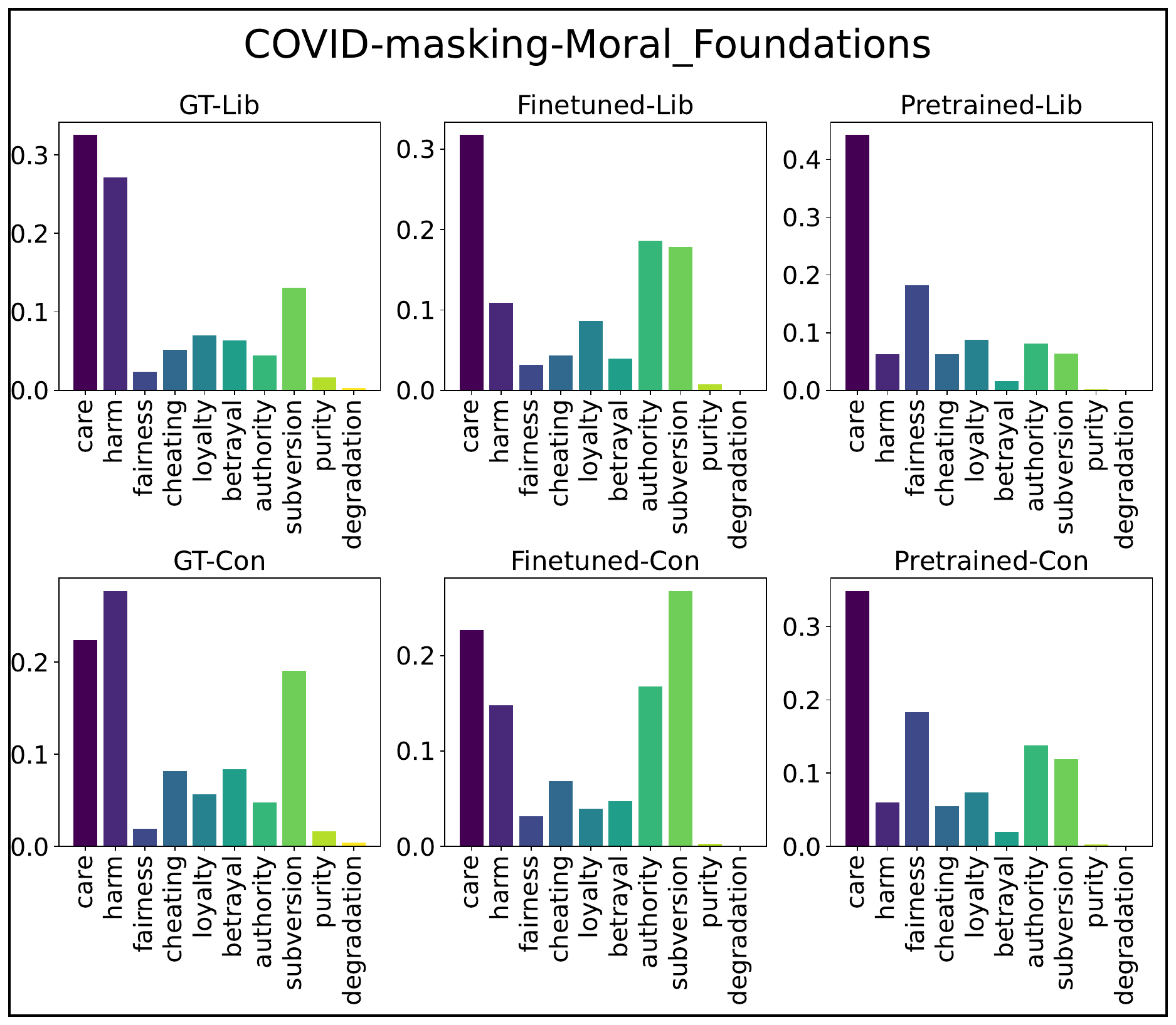}
\end{subfigure}%
\begin{subfigure}{.44\textwidth}
  \centering
  \includegraphics[width=\linewidth]{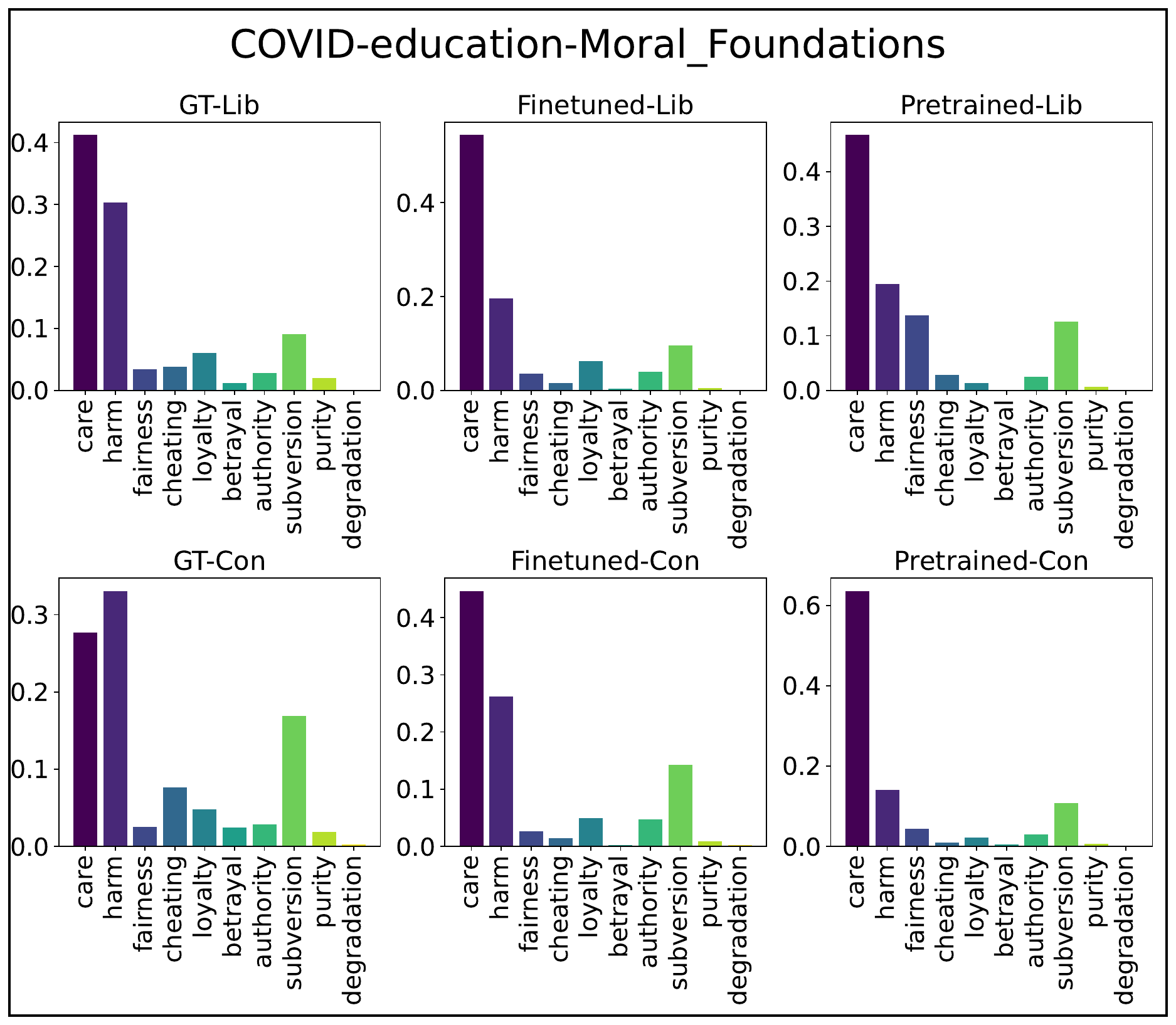}
\end{subfigure}
\\

\begin{subfigure}{.44\textwidth}
  \centering
  \includegraphics[width=\linewidth]{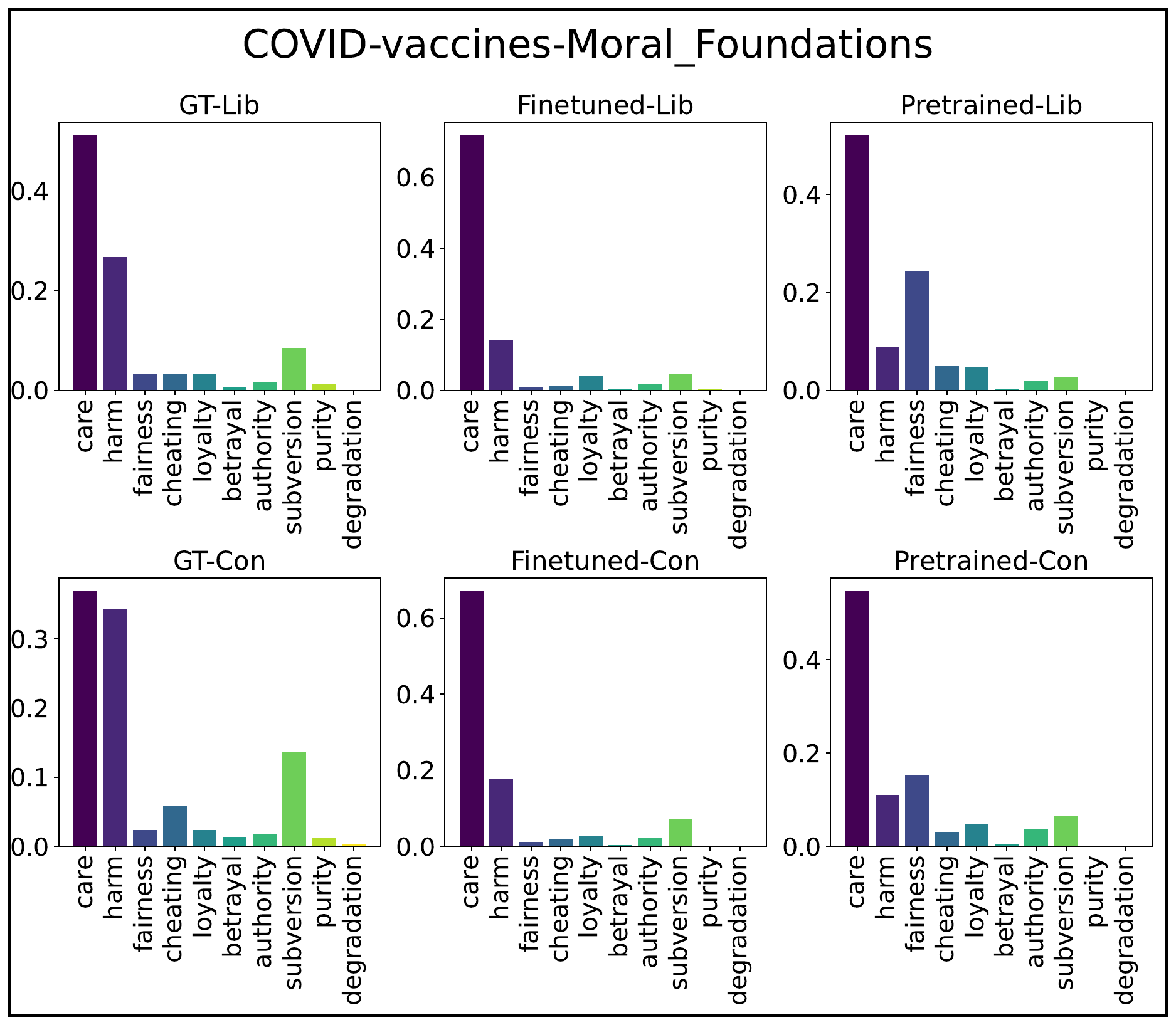}
\end{subfigure}%

\caption{Probability distribution of classes for moral foundation detection in COVID-19 tweets.}

\label{fig:prob-mfs-covid}

\end{figure*}

\begin{figure*}[ht]
\centering

\begin{subfigure}{.44\textwidth}
  \centering
  \includegraphics[width=\linewidth]{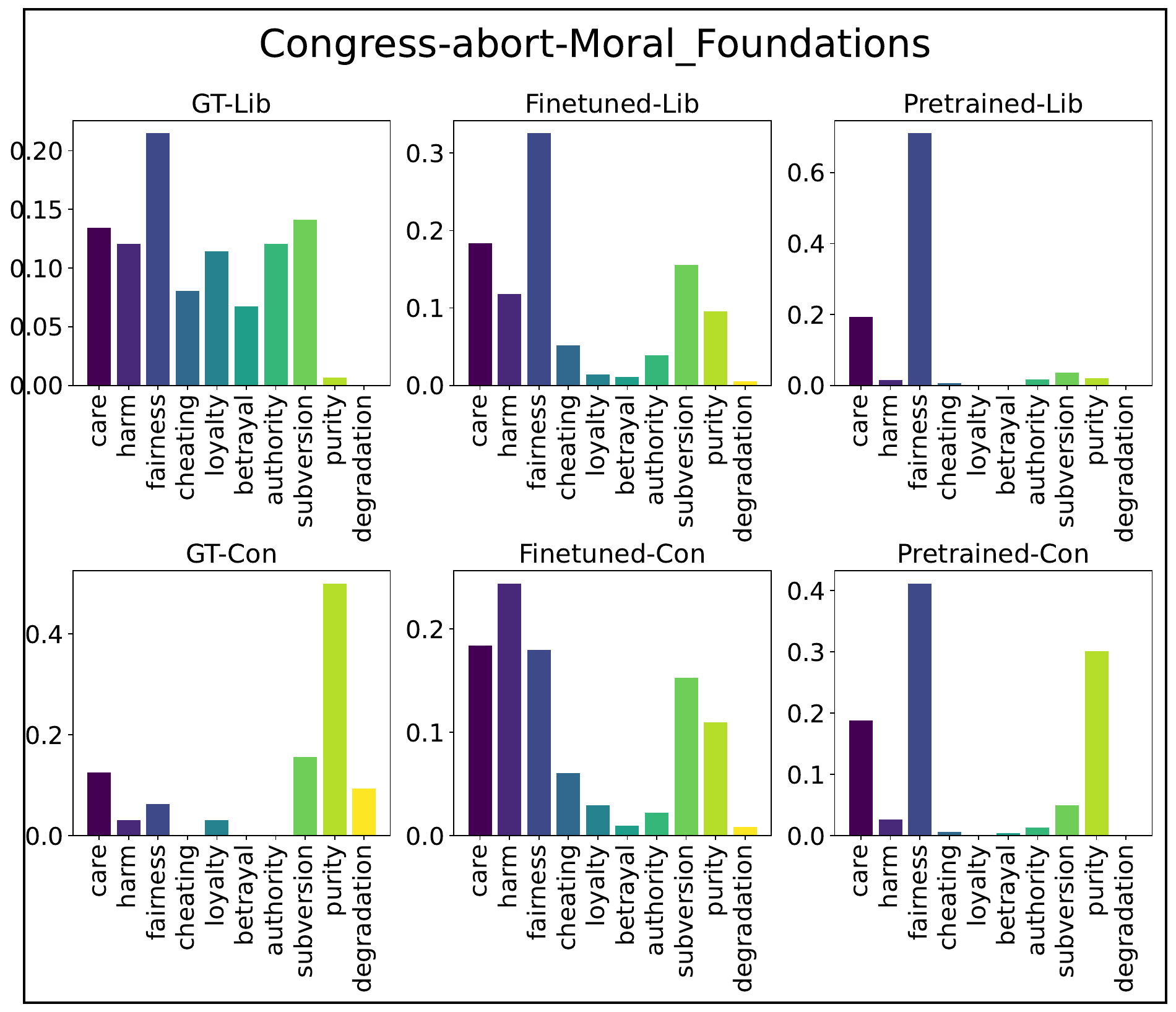}
\end{subfigure}%
\begin{subfigure}{.44\textwidth}
  \centering
  \includegraphics[width=\linewidth]{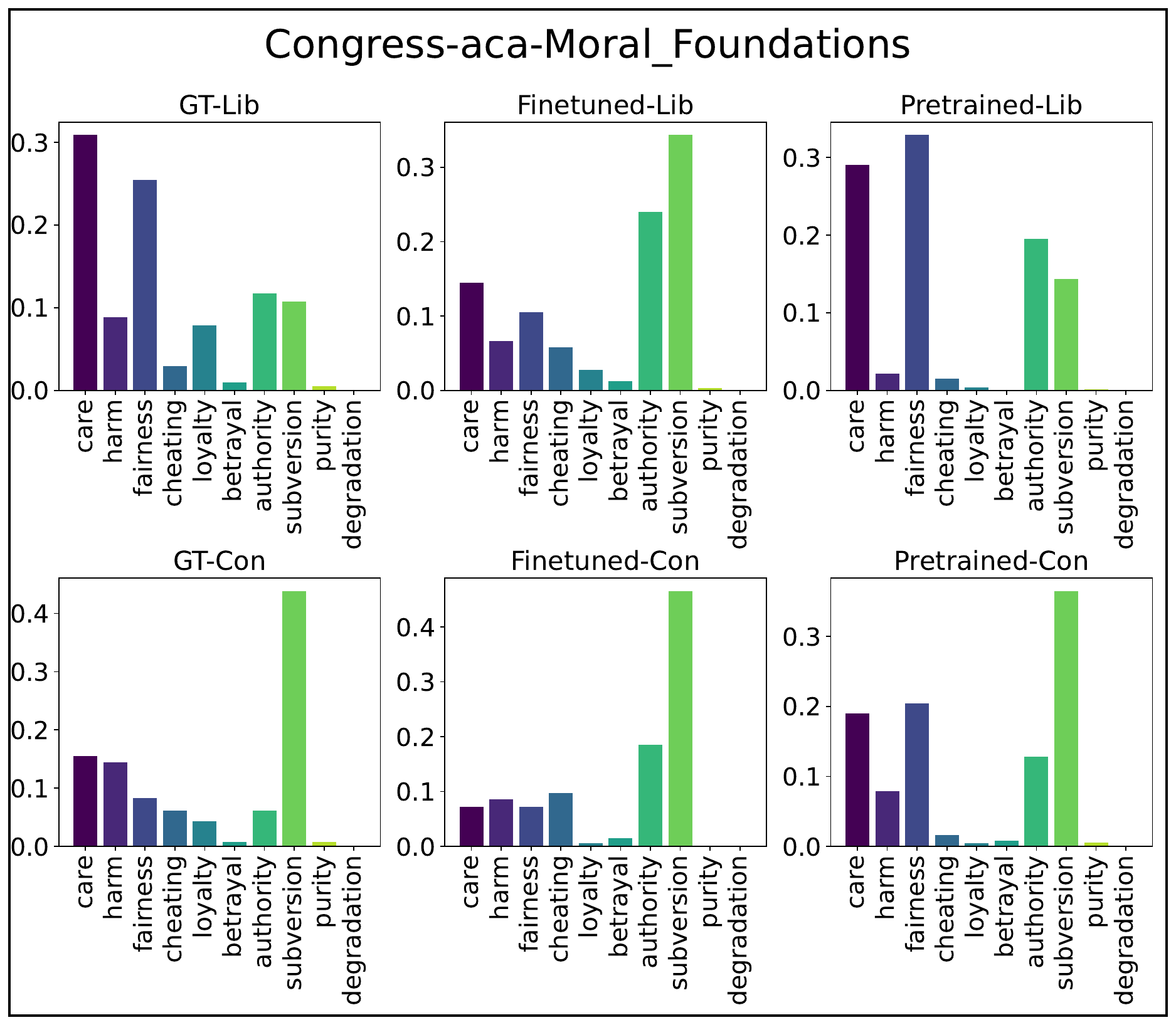}
\end{subfigure}%
\\

\begin{subfigure}{.44\textwidth}
  \centering
  \includegraphics[width=\linewidth]{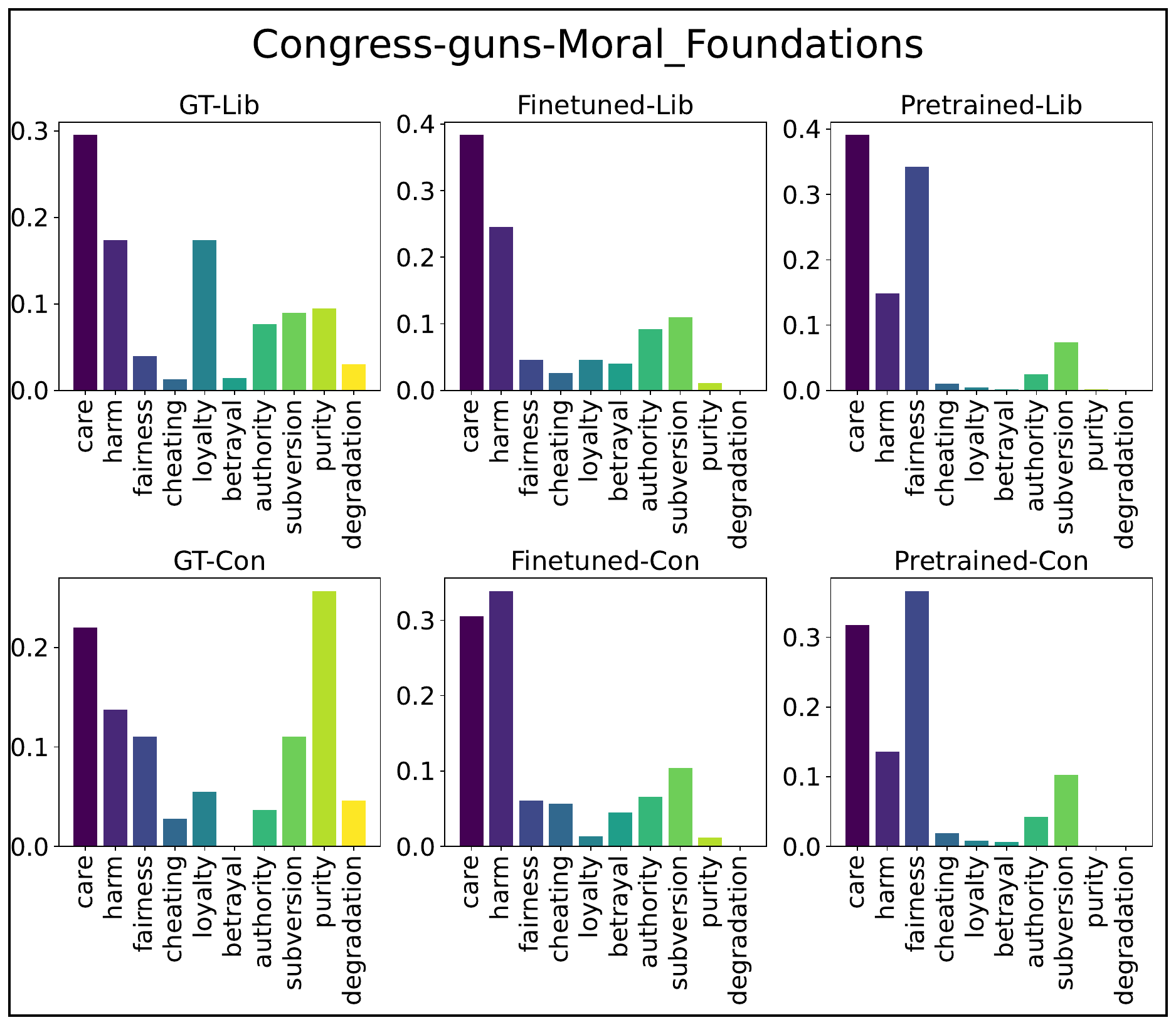}
\end{subfigure}%
\begin{subfigure}{.44\textwidth}
  \centering
  \includegraphics[width=\linewidth]{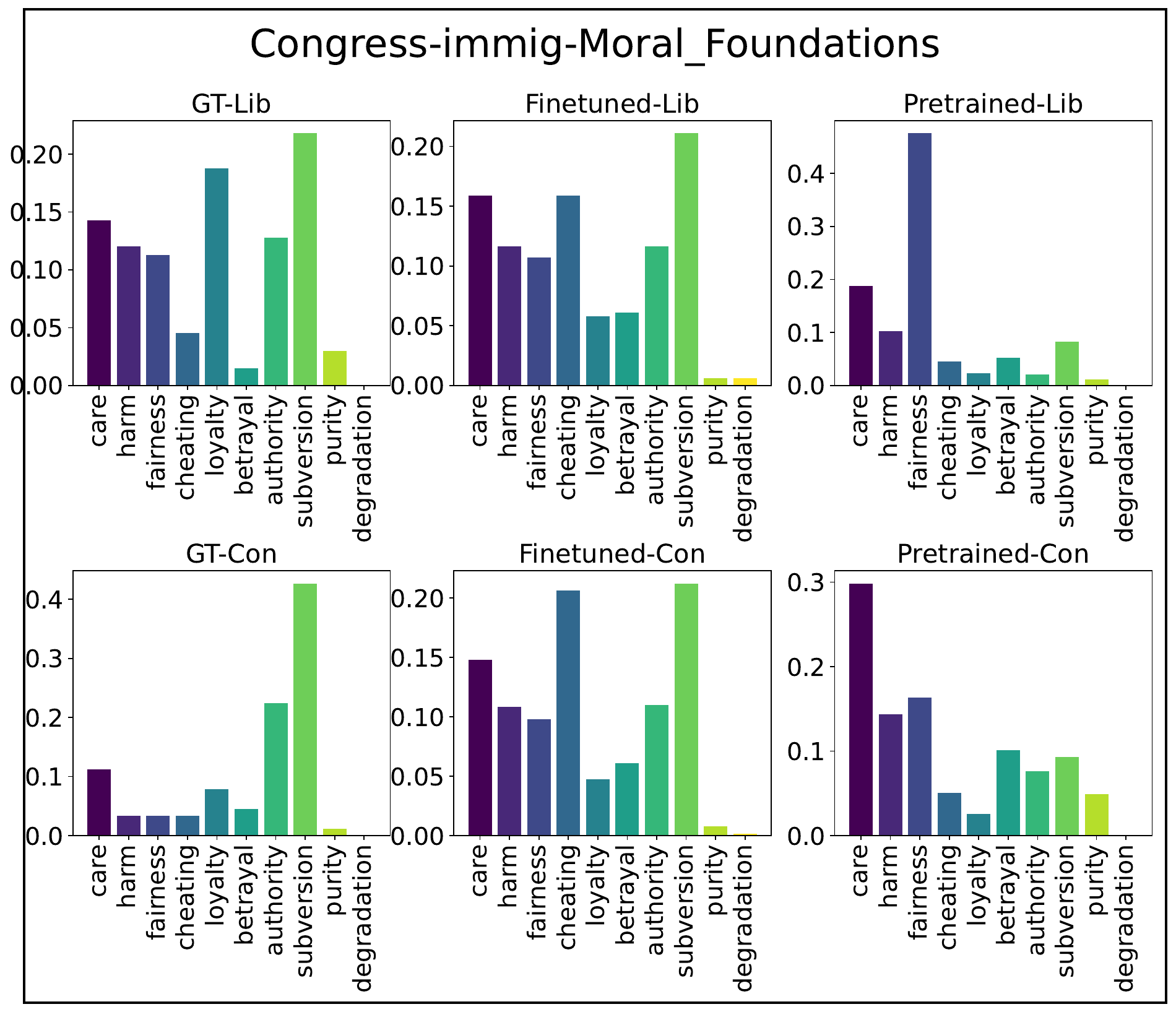}
\end{subfigure}
\\

\begin{subfigure}{.44\textwidth}
  \centering
  \includegraphics[width=\linewidth]{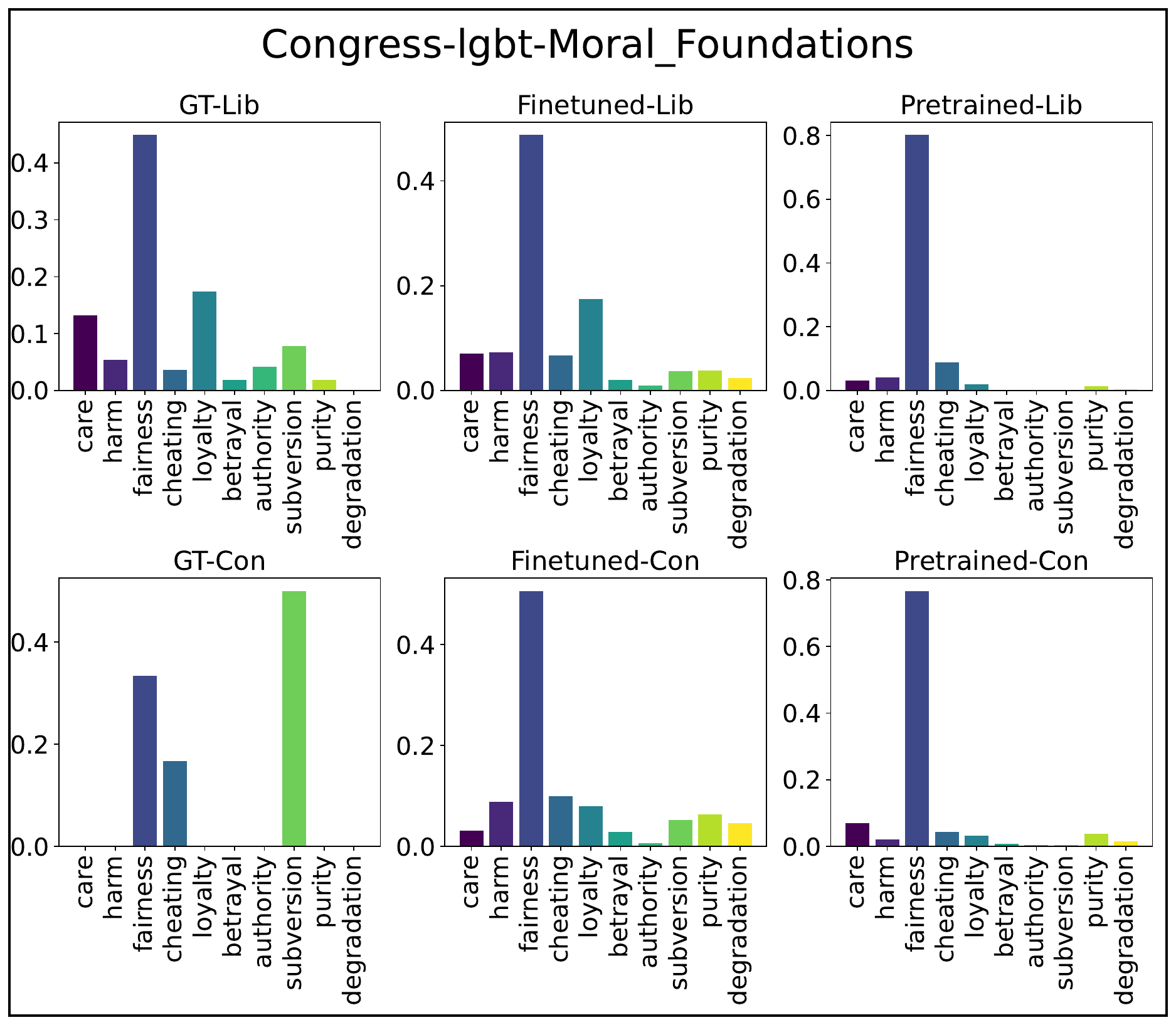}
\end{subfigure}%
\begin{subfigure}{.44\textwidth}
  \centering
  \includegraphics[width=\linewidth]{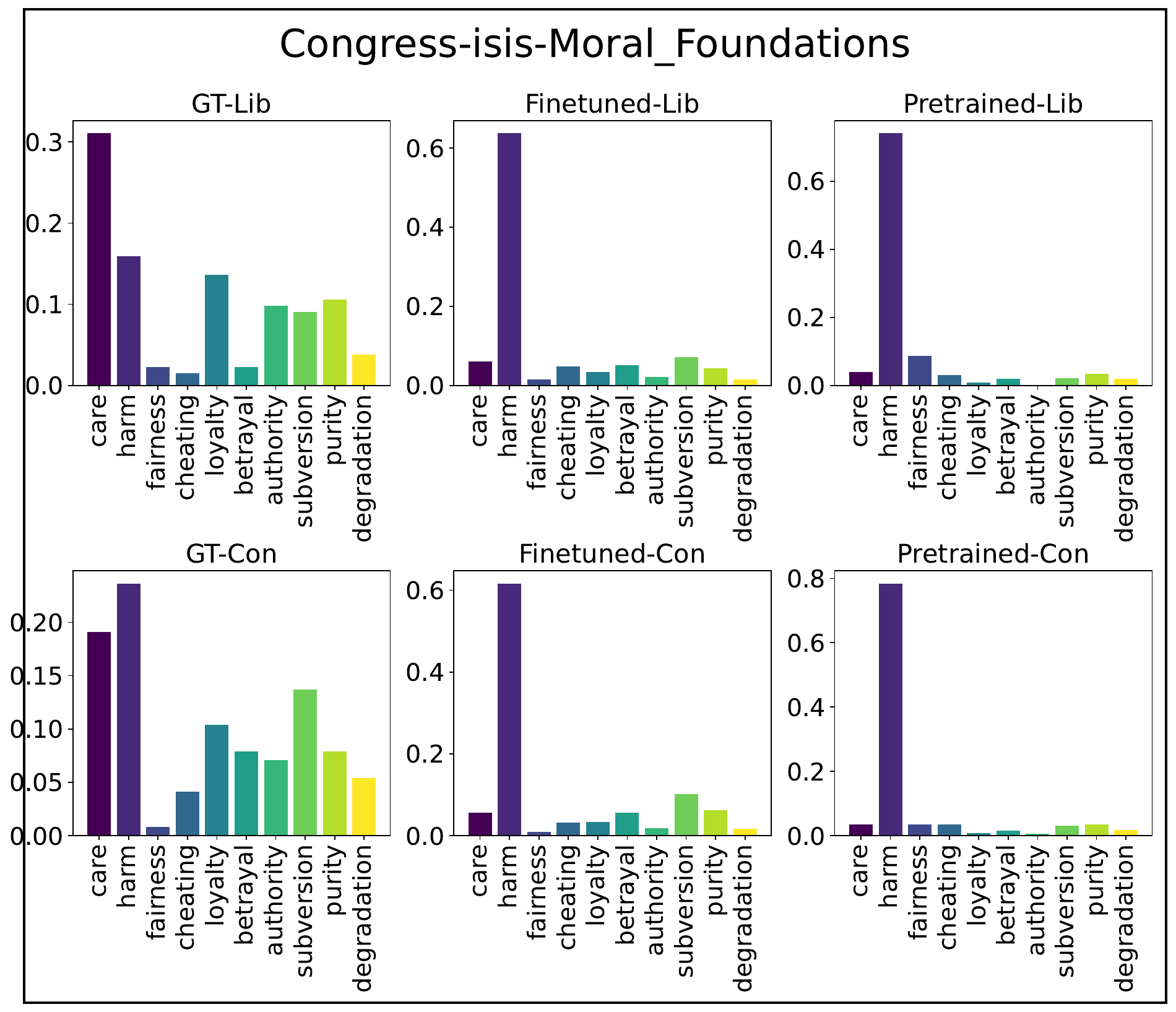}
\end{subfigure}%

\caption{Probability distribution of classes for moral foundation detection in congress tweets.}

\end{figure*}

\label{fig:prob-mfs-congress}

\end{document}